\documentclass{article} % For LaTeX2e
\usepackage[utf8]{inputenc}
\usepackage{arxiv_preprint,times}

\usepackage[table]{xcolor}
\usepackage{appendix}
\usepackage{float}
\usepackage[T1]{fontenc}    % use 8-bit T1 fonts
\usepackage{hyperref}
\hypersetup{colorlinks=true,linkcolor=blue,citecolor=blue,urlcolor=blue}   
\usepackage{etoc}           % local tables of contents (e.g., appendix-only ToC)
\usepackage{graphicx}       % figures
\usepackage{url}            % simple URL typesetting
\usepackage{booktabs}       % professional-quality tables
\usepackage{nicefrac}       % compact symbols for 1/2, etc.
\usepackage{microtype}      % microtypography
\usepackage{amsmath,amssymb,amsfonts}
\usepackage{bbm}
\usepackage{cleveref}
\usepackage{dsfont}
\usepackage{subcaption}
\usepackage{enumitem}
\usepackage{wrapfig}
\usepackage{tabularx}
\usepackage{makecell}
\usepackage{array}

\usepackage[ruled,vlined,linesnumbered]{algorithm2e}
\SetKwInput{KwIn}{Input}

\SetCommentSty{mycommfont}

\usepackage[skins]{tcolorbox}

\definecolor{insightborder}{HTML}{2C3E50} 
\definecolor{insightbg}{HTML}{F4F6F7}     

\newtcolorbox{insightbox}[1][]{
    enhanced,
    boxrule=0pt,
    frame hidden,
    borderline west={3pt}{0pt}{insightborder}, 
    colback=insightbg,
    sharp corners=west, 
    arc=3pt,            
    left=8pt, right=8pt, top=6pt, bottom=6pt, 
    #1
}

\etocsettocdepth{subsection}

\title{SOLAR: A State-Driven Online Learning Rate Scheduler for LLM Pretraining}

\author{%
\resizebox{0.96\textwidth}{!}{\textbf{Qiulin Shang$^{1}$, Binyu Wang$^{2}$, Yongqi Qiao$^{1}$, Songde Rao$^{1}$, Zhoutong Wu$^{1}$, Kun Yuan$^{1}$}} \\
$^{1}$Peking University \quad $^{2}$Nanjing University \\
\texttt{qiulin.shang@stu.pku.edu.cn}
}

\iclrfinalcopy
\begin{document}

\maketitle

\begin{abstract}
Learning-rate (LR) scheduling plays a central role in large language model (LLM) pretraining, yet current practice still relies heavily on hand-crafted heuristics such as Warmup-Cosine-Decay and Warmup-Stable-Decay. Because these schedules are fixed in advance, they cannot adapt to evolving optimization dynamics. Online learned scheduling within the Learning to Optimize (L2O) framework offers a dynamic alternative, but remains brittle at LLM scale due to noisy signals, delayed feedback, and the risk of catastrophic divergence. We propose \textbf{S}tate-driven \textbf{O}nline \textbf{L}earning r\textbf{A}te schedule\textbf{R} (SOLAR), a stabilized framework for reliable online LR adaptation. SOLAR uses a base schedule as a reference and learns bounded, state-dependent residual corrections for individual parameter groups. Each correction re-anchors to the base at every step, allowing the policy to adapt the LR without relearning the warmup--decay profile. A lightweight state representation and progress-aware reward guide online learning, while a Circuit-Breaker restores training after rare unsafe actions. Across autoregressive language-model pretraining, SOLAR improves final perplexity over tuned static schedules and automatic LR tuners for dense models from 60M to 1B, AdamW and Muon, and two MoE settings up to 3B. Matched 130M controls show that adding base anchoring and action bounds improves a global PPO controller from 27.09 to 23.74 final PPL, while group-wise control reaches 22.87 on the same two seeds. A residual policy trained on a 60M proxy can also be frozen and reused at larger dense scales without target PPO updates, remaining effective across a fourfold base-LR range. These results establish SOLAR as a practical learned LR controller for LLM pretraining.
\end{abstract}

\section{Introduction}
\label{intro}
Despite decades of progress in optimization, the learning-rate schedule for pretraining billion-parameter large language models (LLMs) is still set by hand and fixed before training begins. In practice, hand-crafted schedules such as Warmup-Cosine-Decay~(Cosine)~\citep{loshchilov2017sgdrstochasticgradientdescent} and Warmup-Stable-Decay (WSD)~\citep{hu2024minicpmunveilingpotentialsmall} remain the default due to their simplicity, reliability, and ease of deployment~\citep{LLaMA, touvron2023llama2openfoundation}. Yet they impose a fundamental limitation: once training begins, the schedule is entirely predetermined and cannot adapt to the evolving optimization landscape, including shifts in loss dynamics, gradient norms, and parameter magnitudes. Adaptive optimizers do not resolve this limitation: although methods such as Adam~\citep{kingma2014adam} and Muon~\citep{jordan2024muon} rescale updates at the individual parameter level, the global learning rate that governs the overall update magnitude is still specified in advance.

A natural alternative is \emph{online learning-rate scheduling}, where the learning rate is adjusted on the fly according to current and historical training signals~\citep{baydin2018onlinelearningrateadaptation, autoLRS, learningstep}. This approach aligns closely with the broader Learning-to-Optimize (L2O) paradigm, as it requires the scheduler to dynamically adapt based on the unfolding optimization state. In large-scale LLM pretraining, however, online scheduling is notoriously difficult to operationalize. The available optimization signals, such as loss trends and gradient norms, are noisy---that is, they are high-variance, non-stationary, and often only weakly informative at the level of individual steps when optimizing a nonconvex objective with stochastic methods~\citep{PaLM, Spam, mccandlish2018empiricalmodellargebatchtraining}. A suboptimal scheduling decision may not trigger an immediate failure yet can silently degrade progress over a long horizon; conversely, an overly aggressive adjustment can cause abrupt instability within a few steps~\citep{takase2025spikemorestabilizingpretraining}. These delayed and asymmetric failure modes make learned exploration of the learning rate particularly brittle in realistic pretraining runs. As a result, although learned schedulers have shown promise in smaller settings, they have not yet been widely demonstrated as reliable solutions in realistic LLM pretraining settings. This gap motivates the search for an online scheduler that is not only adaptive but also robust enough to operate in such settings.

\begin{figure}[t]
    \centering
    \includegraphics[width=0.9\linewidth]{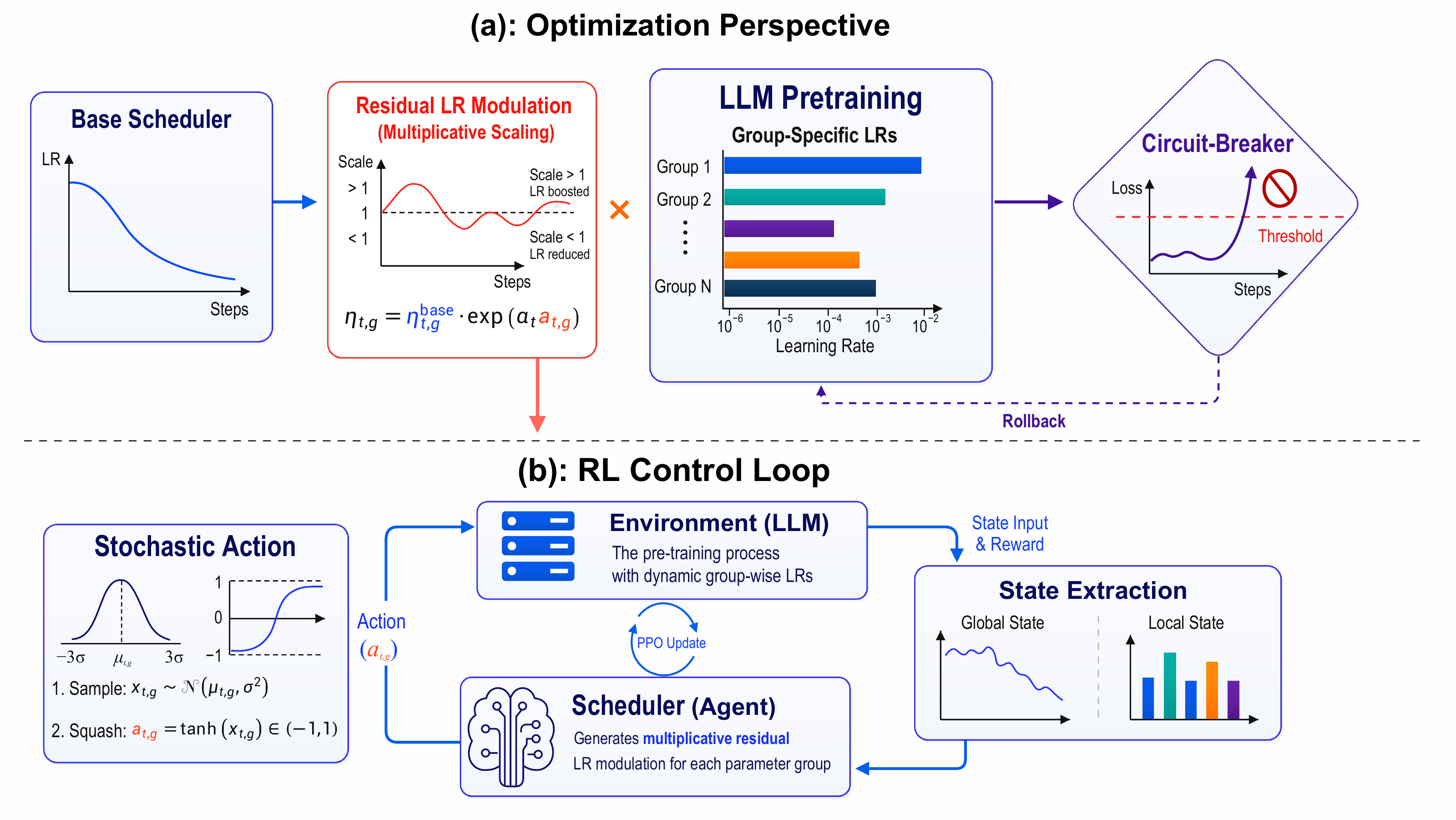}
    \caption{\small Overview of SOLAR. Panel (a) presents the optimization perspective: SOLAR augments a base scheduler with residual learning-rate modulation, applies bounded group-wise LR actions to the LLM pretraining process, and uses an automated Circuit-Breaker to roll back unsafe trajectories when severe loss spikes are detected. Panel (b) illustrates the RL control loop: SOLAR constructs lightweight global and local state features from the current training dynamics, outputs parameter-group-wise residual actions, receives delayed reward feedback after the optimizer update, and improves the scheduler through PPO updates.}
    \label{fig:overview}
    \vspace{-6mm}
\end{figure}

\begin{wrapfigure}{r}{0.38\textwidth}
    \centering
    \vspace{-4pt}
    \includegraphics[width=\linewidth]{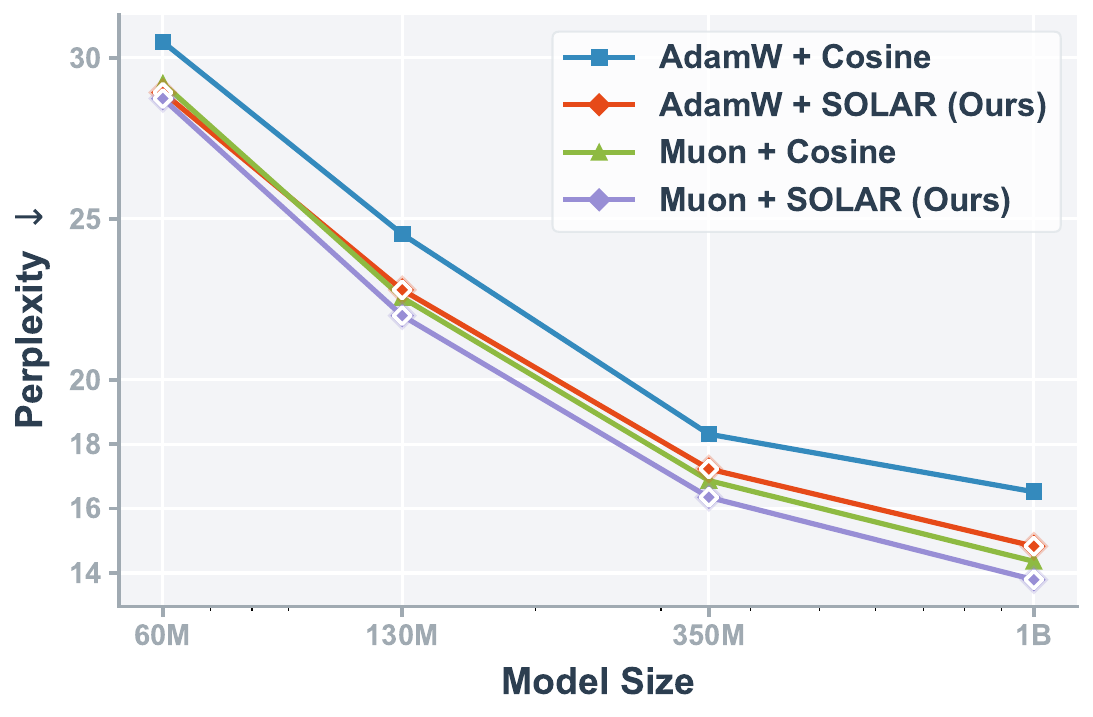}
    \vspace{-6pt}
    \caption{\small Pretraining performance on C4 across model scales. SOLAR consistently achieves the lowest perplexity, outperforming strong baselines in both AdamW and Muon families.}
    \label{fig:scaling}
    \vspace{-8pt}
\end{wrapfigure}

In this work, we introduce \textbf{SOLAR} (\textbf{S}tate-driven \textbf{O}nline \textbf{L}earning r\textbf{A}te schedule\textbf{R}), a reinforcement learning (RL)-based L2O framework for online learning-rate scheduling in LLM pretraining. Rather than generating the full LR trajectory, SOLAR uses a base schedule as a reference and learns bounded, state-dependent residual corrections for individual parameter groups. The base supplies the warmup and decay profile, while the policy adapts to the realized optimization state. Each correction re-anchors to the current base LR, preventing exploratory errors from compounding into a new schedule. As illustrated in Figure~\ref{fig:overview}, SOLAR maps lightweight training-state features to parameter-group-wise LR adjustments online. Its stochastic residual actions support fine-grained exploration, and action-scale warmup limits early perturbations. A progress-aware reward combines immediate improvement, longer-term trend, and group stability. A threshold-based Circuit-Breaker detects severe loss spikes and restores the latest safe checkpoint.

We evaluate SOLAR on autoregressive language modeling over C4~\citep{C4} for Llama~2~\citep{touvron2023llama2openfoundation} models ranging from 60M to 1B parameters, using AdamW~\citep{loshchilov2017decoupled} and Muon. For the 1B model, SOLAR reduces perplexity (PPL) by over 10\% relative to AdamW with Cosine and by about 4\% relative to Muon with Cosine (Figure~\ref{fig:scaling}). Repeated runs preserve the gains at every repeated scale, including two paired 1B AdamW runs. We further validate SOLAR on pretraining a 1B-parameter Qwen2-MoE~\citep{qwen2technicalreport} model on The Pile~\citep{pile} under AdamW. SOLAR improves final perplexity from 9.61 to 9.34 while reducing early-stage optimization volatility (Figure~\ref{fig:moe_pretrain}); a 3B MoE run improves from 10.73 to 10.38 (Appendix~\ref{app:megatron_3b_moe}). Matched 130M controls clarify why the method works: a recursive global PPO controller ends at 27.09 PPL, base anchoring and action bounds improve it to 23.74, and group-wise control reaches 22.87 on the same two seeds. A residual policy trained exclusively on a 60M proxy can also be frozen and reused at larger dense scales without target PPO updates or SOLAR-specific retuning; it remains effective across $0.5\times$--$2\times$ base LRs.

Overall, our main contributions are:
\begin{itemize}[leftmargin=1.4em,itemsep=0.25em,topsep=0.25em]
\item \textbf{Methodological Design.} We propose SOLAR, an RL-based framework for online LR scheduling in LLM pretraining. SOLAR keeps the base LR profile outside the learned controller and learns bounded, state-dependent group-wise corrections that re-anchor at every step. A progress-aware reward and Circuit-Breaker support policy learning throughout a full pretraining run.

\item \textbf{Empirical Demonstration.} SOLAR improves final PPL over tuned static schedules and automatic LR tuners across dense and MoE models under AdamW and Muon, with repeated gains through 1B. On matched seeds, controls show that base anchoring and action bounds improve a global PPO controller by 3.35 PPL, while group-wise control contributes another 0.87 PPL.

\item \textbf{Practical Scalability.} A residual policy acquired through full-length online runs on a 60M proxy can be frozen and reused at larger dense scales without target PPO updates or SOLAR-specific retuning, while remaining effective across a fourfold base-LR range.
\end{itemize}

\section{Related Work}
\label{sec:related_work}

\paragraph{Learning Rate Scheduling and Hyperparameter Optimization.}
Preset schedules encode the open-loop training profile. Cosine~\citep{loshchilov2017sgdrstochasticgradientdescent,LLaMA}, WSD~\citep{hu2024minicpmunveilingpotentialsmall}, CLR~\citep{clr}, and one-cycle schedules~\citep{smith2018superconvergencefasttrainingneural} choose warmup and decay before observing the realized run. Blockwise LR~\citep{blockwise} assigns different rates to transformer module types but remains preset. Hyperparameter-optimization methods search this profile: Bayesian optimization~\citep{snoek2012practicalbayesianoptimizationmachine} and population-based training~\citep{jaderberg2017populationbasedtrainingneural} use multiple trials, while AutoLRS~\citep{autoLRS} evaluates short candidate segments online. MECHANIC~\citep{mechanic} instead adapts a global multiplier. SOLAR keeps the base profile outside the learned controller and learns closed-loop corrections within the live target run.

\paragraph{Learned Learning-Rate Controllers.}
Learned LR control has a substantial lineage. Early work uses reinforcement learning to select optimization hyperparameters or step sizes~\citep{hansen2016deepq,learningstep,xu2017rlcontrol}. \citet{xu2019adaptive} train a PPO controller from past training histories and transfer it across image tasks. GNS~\citep{xiong2022gns} encodes layer states with a graph network and learns a global schedule across target episodes. \citet{subramanian2023learned} study PPO scheduling on MNIST and CIFAR, while GANNO~\citep{tessera2023ganno} uses layer-wise agents and absolute LR actions. These studies establish RL scheduling, state conditioning, transfer, and layer-wise control. SOLAR addresses a different operating constraint: a single full autoregressive LLM pretraining trajectory must both train the controller and remain competitive. The base schedule supplies the open-loop profile; the policy learns only bounded residual corrections that re-anchor every step. To our knowledge, SOLAR is the first learned LR scheduler shown to train online throughout the same full LLM pretraining run that it improves. Appendix~\ref{app:learned_lr_positioning} compares the learned objects and acquisition protocols, and Appendix~\ref{app:novelty_controls} evaluates matched controllers in the LLM setting.

\paragraph{Adaptive and Schedule-Free Optimizers.}
Adaptive optimizers modify update magnitudes from gradient history. AdaGrad~\citep{duchi2011adaptive}, RMSProp~\citep{tieleman2012rmsprop}, AdamW~\citep{loshchilov2017decoupled}, Adafactor~\citep{shazeer2018adafactor}, Lion~\citep{chen2023symbolicdiscoveryoptimizationalgorithms_lion}, and Muon~\citep{jordan2024muon} differ in their internal update geometry, but still accept an external LR trajectory. D-Adaptation~\citep{defazio2023learningratefreelearningdadaptation}, Prodigy~\citep{mishchenko2024prodigyexpeditiouslyadaptiveparameterfree}, and Schedule-Free optimization~\citep{defazio2024roadscheduled} reduce that dependence through optimizer-side scalar adaptation or iterate averaging. SOLAR leaves the optimizer update unchanged and controls the LR across parameter groups from observed training states. The AdamW and Muon experiments test whether this scheduler-level control remains useful across distinct optimizer families.

\paragraph{Learning to Optimize.}
Learning to Optimize (L2O) replaces hand-designed update rules with learned ones~\citep{chen2021learningoptimizeprimerbenchmark}. Learned optimizers map gradients to updates~(\mbox{\citealp{andrychowicz2016learninglearngradientdescent}}; \mbox{\citealp{li2016learningoptimize}}; \mbox{\citealp{ravi2017optimization}}); hierarchical models~\citep{wichrowska2017learnedoptimizersscalegeneralize}, large-scale meta-training~\citep{metz2022velotrainingversatilelearned}, and constrained update spaces~\citep{liu2023constitutingmathematicalstructureslearning,he2024mathematicsinspiredlearningtooptimizeframeworkdecentralized} improve their reach. SOLAR learns a lower-dimensional object. The base optimizer retains its update rule, and the policy controls a bounded residual around an external schedule. This restriction makes online acquisition feasible inside the target pretraining run.

\section{Preliminaries}
\label{sec:preliminaries}

\textbf{Notation.} We consider LLM pretraining as minimizing a loss $\mathcal{L}(w)$ over parameters $w\in\mathbb{R}^d$ with stochastic gradients $g_t=\nabla\mathcal{L}(w_t)$. The parameters are partitioned into $G$ controlled groups, each assigned a learning rate $\eta_{t,g}>0$. SOLAR supplies $\{\eta_{t,g}\}_{g=1}^{G}$ to a standard optimizer (e.g., AdamW, Muon), which internally produces the next iterate $w_{t+1}$ during optimization.

\textbf{Learning Rate Scheduling as a Sequential Decision Problem.}
We formulate online LR scheduling as a sequential decision process. At step $t$, the scheduler observes state $s_t=\{s_{t,g}\}_{g=1}^{G}$, outputs the group-wise LR action vector $a_t=\{a_{t,g}\}_{g=1}^{G}$, and receives rewards $r_{t+1}=\{r_{t+1,g}\}_{g=1}^{G}$. The full action vector is applied to one LLM optimizer step, which jointly determines the next training state and the group-wise rewards.

\textbf{Proximal Policy Optimization (PPO).}
We train the scheduler with PPO~\citep{schulman2017proximalpolicyoptimizationalgorithms} using a parameter-shared independent update across groups. Each controlled group contributes a likelihood ratio paired with its own advantage. The clipped terms are averaged to update the shared actor--critic. Appendix~\ref{app:action_ppo_impl} gives the exact objective, network architecture, and hyperparameters.
\section{SOLAR: State-Driven Online Learning Rate Scheduler}
\label{sec:solar}

\begin{insightbox}
\textbf{SOLAR at a Glance.} A base schedule specifies warmup and decay. SOLAR learns the missing closed-loop correction from the current run. It maps global and parameter-group states to bounded residual actions that re-anchor to the base every step. A progress-aware reward trains this policy online, and a Circuit-Breaker restores the last safe state after a severe loss spike.
\end{insightbox}

\subsection{Lightweight State Representation}
\label{subsec:state}

SOLAR's state design follows a minimal-variance principle: it relies on a compact set of low-variance optimization signals that are informative yet cheap to compute, avoiding high-dimensional or optimizer-internal statistics. In the dense implementation, each controlled parameter group corresponds to one trainable parameter tensor; the MoE implementations follow the module partition of their respective codebases. SOLAR assigns one learning-rate multiplier per group. To minimize the overhead of online scheduling, SOLAR adopts a lightweight state representation built on a small set of universal optimization signals: loss statistics, gradient norms, and parameter norms.

\paragraph{State representation.}
For controlled parameter group \(g\), the state is defined as \(s_{t,g} = [s_t^{\mathrm{global}};\; s_{t,g}^{\mathrm{local}}]\). The global state summarizes the macro training phase and loss dynamics:
\begin{equation}
s_t^{\mathrm{global}}
=
\left[
\tau_t,\;
\log L_t,\;
\nu_t,\;
\Delta \mathrm{EMA}_t
\right],
\label{eq:global_state}
\end{equation}
where \(\tau_t\) is the normalized training progress, \(L_t\) is the current loss, \(\nu_t\) measures short-window loss fluctuation, and \(\Delta \mathrm{EMA}_t\) captures the discrepancy between short-term and long-term exponential moving averages of the loss. The local state summarizes controlled-parameter-group-level context:
\begin{equation}
s_{t,g}^{\mathrm{local}}
=
\left[
\log \eta^{\mathrm{base}}_{t,g},\;
\log \|\nabla_{t,g}\|,\;
a_{t-1,g},\;
d_g,\;
\log \|w_{t,g}\|,\;
\Delta \log \|\nabla_{t,g}\|
\right],
\label{eq:local_state}
\end{equation}
where \(\nabla_{t,g}\) is the current mini-batch gradient, \(a_{t-1,g}\) is the previous scheduling action, \(d_g\) is the depth index, \(w_{t,g}\) is the trainable tensor, and \(\Delta \log \|\nabla_{t,g}\|\) measures recent changes in gradient magnitude. This compact representation keeps the scheduler lightweight and decoupled from optimizer-internal statistics. Detailed mathematical formulations for all state features, including moving average decay rates and numerical stability constants, are provided in Appendix~\ref{app:state_impl}.

\subsection{Action Space: Stochastic Group-wise Residual Modulation}
\label{subsec:action}

Given the observed state, a conventional action design is to let the policy directly output an absolute learning rate deterministically. However, accurately predicting a single optimal learning rate at each step is inherently difficult in noisy, large-scale, and highly non-stationary LLM pretraining. SOLAR addresses this challenge through two coupled design choices under practical constraints.

\paragraph{Stochastic group-wise residual actions.}
Instead of predicting a deterministic scalar learning rate, SOLAR samples a joint collection of one-dimensional bounded residual learning-rate actions across controlled parameter groups from a factorized squashed Gaussian policy. The joint policy factorizes across groups, with one squashed Gaussian factor per controlled parameter group. For each controlled parameter group $g$, the scheduler outputs a mean $\mu_{t,g}$, while the policy uses a single learnable scalar standard deviation $\sigma$ shared across all controlled parameter groups. We sample a latent variable, apply the hyperbolic tangent, and clip only the executed action for numerical safety:
\[
u_{t,g}\sim\mathcal N(\mu_{t,g},\sigma^2),\qquad
\widetilde a_{t,g}=\tanh(u_{t,g}),\qquad
a_{t,g}=\operatorname{clip}(\widetilde a_{t,g},-1+10^{-4},1-10^{-4}).
\]

The stochastic policy explores group-specific corrections, while the bounded action limits the damage of any single decision. We use group-specific means \(\mu_{t,g}\) and one learned scalar variance shared across groups; Appendix~\ref{app:action_ppo_impl} gives the implementation.

\paragraph{Residual LR modulation.}
A base schedule encodes the coarse LR profile. SOLAR keeps that profile outside the learned object and controls a residual:
\begin{equation}
\eta_{t,g}
=
\eta^{\mathrm{base}}_{t,g}\cdot \exp(\alpha_t a_{t,g}),
\label{eq:residual_lr}
\end{equation}
where \(\alpha_t \ge 0\) controls the residual range. We linearly warm up \(\alpha_t\) from 0 to \(\alpha\) during early training. Because each multiplier is applied to the current base LR, the action cannot recursively redefine later LRs. The policy can focus on state-dependent corrections instead of relearning warmup and decay.

\subsection{Reward Design for Sustained Optimization}
\label{subsec:reward}

After action \(a_t\) and the optimizer update, the next ordinary forward/backward pass produces \(L_{t+1}\) and \(\nabla_{t+1}\). SOLAR reuses these detached quantities and adds no language-model forward or backward pass. A one-step loss reward is too noisy to describe sustained progress, so the reward combines immediate improvement, an EMA trend, and group-wise stability:

\begin{equation}
\label{eq:reward_rt}
    r_{t+1,g} = r_{t+1}^{\mathrm{perf}} + r_{t+1}^{\mathrm{trend}} - p_{t+1,g}^{\mathrm{stab}},
\end{equation}
where \(r_{t+1}^{\mathrm{perf}}\) rewards step-wise loss reduction, \(r_{t+1}^{\mathrm{trend}}\)
encourages improvements over a longer EMA horizon, and \(p_{t+1,g}^{\mathrm{stab}}\) is a signed,
controlled-group-wise stability shaping term (Appendix~\ref{app:reward_cb_impl}). It compares each
group's post-update gradient norm with its recent trend, crediting gradients that settle below the
trend and penalizing growth above it. The two progress terms remain tied to the global
language-modeling objective.

\subsection{Safety and Recovery: Circuit-Breaker Mechanism}
\label{subsec:circuit_breaker}

Rare loss spikes can occur during LLM pretraining~\citep{PaLM, zhang2022optopenpretrainedtransformer, zeng2023glm130bopenbilingualpretrained} and may invalidate online exploration. SOLAR therefore adds a global emergency trigger that terminates unsafe rollouts and restores the latest safe checkpoint. Specifically, we augment the reward with an explicit instability penalty:

\begin{equation}
r^{\mathrm{final}}_{t+1,g}
=
r_{t+1,g} - \lambda_{\mathrm{cb}}\psi_{t+1},
\label{eq:cb_reward}
\end{equation}
where \(\lambda_{\mathrm{cb}}\) is a large penalty coefficient, and \(\psi_{t+1}\) is a binary instability indicator defined as
\begin{equation}
\psi_{t+1}
=
\mathbbm{1}\!\left[
L_{t+1} > \kappa_L\cdot L^{\mathrm{ema}}_{t+1}
\right],
\label{eq:psi_indicator}
\end{equation}
with \(\kappa_L\) denoting a safety threshold, \(L_{t+1}\) the post-update training loss, and \(L^{\mathrm{ema}}_{t+1}\) its exponential moving average. Unlike the per-group stability shaping term in \eqref{eq:reward_rt}, the Circuit-Breaker acts as a global emergency signal triggered only by severe loss spikes. When \(\psi_{t+1}=1\), the current rollout is immediately terminated and the scheduler performs a PPO update on the truncated trajectory. The training loop then restores the model and optimizer states from the latest safe checkpoint while retaining the updated PPO state. This prevents post-failure samples from contaminating subsequent policy updates and provides an explicit failure signal that encourages the scheduler to internalize stability constraints.

\subsection{Implementation Details and Overhead}
\label{subsec:implementation}

SOLAR uses a two-layer MLP actor--critic trained with PPO; Appendix~\ref{app:action_ppo_impl} gives the architecture. Appendix~\ref{app:runtime_overhead} separates steady-state step time from full-run wall-clock accounting.

\begin{algorithm}[t]
\small
\caption{State-Driven Online Learning Rate Scheduler (SOLAR)}
\label{alg:solar}

\KwIn{\parbox[t]{0.88\linewidth}{Base scheduler \(\{\eta^{\mathrm{base}}_{t,g}\}\), optimizer \(\mathcal{O}\), policy \(\pi_\theta\), value network \(V_\phi\), total steps \(T\)}}

\For{\(t = 0, \dots, T-1\)}{
    \tcp{1. State Construction}
    Construct scheduler states \(\{s_{t,g}\}_{g=1}^G\)\;
    
    \tcp{2. Per-Group LR Action Sampling}
    Sample \(\{u_{t,g}\}_{g=1}^G\), form \(\{a_{t,g}\}_{g=1}^G\), and record \(\{\ell^{\mathrm{old}}_{t,g}\}_{g=1}^G\)\;
    Set learning rates \(\eta_{t,g} = \eta^{\mathrm{base}}_{t,g} \exp(\alpha_t a_{t,g})\) for all \(g\)\;
    
    \tcp{3. Environment Step}
    Execute one optimizer step with \(\{\eta_{t,g}\}_{g=1}^G\)\;
    Compute group-wise final rewards \(\{r^{\mathrm{final}}_{t+1,g}\}_{g=1}^G\) and instability indicator \(\psi_{t+1}\)\;
    Store \((s_t,u_t,a_t,\ell_t^{\mathrm{old}},r^{\mathrm{final}}_{t+1},s_{t+1})\), retaining the group axis, in the rollout buffer\;

    \tcp{4. Policy Update \& Safety Net}
    \If{the PPO update condition is satisfied or \(\psi_{t+1}=1\)}{
        Update \(\pi_\theta\) and \(V_\phi\) using PPO\;
    }
    \If{\(\psi_{t+1} = 1\)}{
        \tcp{Circuit-Breaker Triggered}
        Emit abort signal\;
        The outer training loop restores model and optimizer states from the latest safe checkpoint\;
    }
}
\end{algorithm}

\section{Experiments}
\label{sec:experiments}

We ask four questions: Does SOLAR improve final pretraining outcomes? Which control choices make online learning effective? Does the residual policy transfer across scales and base schedules? What computational cost does the controller add?

\subsection{Main Pretraining Results}
\label{sec:exp_online_effectiveness}

\subsubsection{Dense Llama~2 Pretraining across Optimizers}

We evaluate on C4~\citep{C4} across four Llama~2~\citep{touvron2023llama2openfoundation} scales (60M to 1B) with AdamW and Muon, following the setup of~\citet{zhao2024galorememoryefficientllmtraining}. Static schedules, online tuners, and schedule-free baselines share the model, data, optimizer, and training budget. Table~\ref{tab:main_results} reports validation PPL at the final training update. SOLAR-specific hyperparameters are selected once at 60M and then kept fixed. Appendix~\ref{app:tuning_protocol} gives the complete selection and evaluation protocol.

\begin{table}[t]
\centering
\caption{\small Seed-52 final-checkpoint validation perplexity on C4. SOLAR-online acquires its policy within the target run; SOLAR-frozen reuses a policy acquired from full-length online runs at 60M, with no target PPO updates. AvgLR Replay replays SOLAR's step-wise mean LR. $\dagger$ marks official AdamW-based implementations. Multi-seed results appear in Appendix~\ref{app:seed_results}.}
\label{tab:main_results}

\begin{tabular}{lcccc}
\toprule

& \multicolumn{4}{c}{\textbf{Model Scale (Validation Perplexity $\downarrow$)}} \\

\cmidrule(lr){2-5}
\textbf{Method} & \textbf{60M} & \textbf{130M} & \textbf{350M} & \textbf{1B} \\
\midrule

\multicolumn{5}{l}{\textit{\textbf{Base Optimizer: AdamW}}} \\
\midrule
\quad Cosine       & 30.49 & 24.52 & 18.31 & 16.52 \\
\quad AvgLR Replay & 30.68 & 25.61 & 21.69 & 20.06 \\
\quad WSD                   & 29.80 & 23.96 & 18.75 & 16.29 \\
\quad CLR                   & 31.78 & 28.14 & 21.43 & 20.22 \\
\quad $\dagger$Blockwise LR   & 31.06 & 24.38 & 18.56 & 16.74 \\
\quad AutoLRS              & 31.46 & 25.02 & 20.07 & 19.08 \\
\quad MECHANIC              & 32.14 & 26.95 & 20.60 & 19.25 \\
\quad $\dagger$Prodigy               & 45.27 & 27.51 & 22.14 & 20.75 \\
\quad $\dagger$Schedule-Free   & 30.28 & 24.49 & 18.14 & 16.45 \\
\rowcolor{gray!10} \quad \textbf{SOLAR (online)} & \textbf{28.92} & \textbf{22.79} & \textbf{17.23} & \textbf{14.83} \\
\rowcolor{gray!10} \quad \textbf{SOLAR (frozen)} & -- & \textbf{22.41} & \textbf{17.37} & \textbf{14.63} \\
\midrule
\multicolumn{5}{l}{\textit{\textbf{Base Optimizer: Muon}}} \\
\midrule
\quad Cosine       & 29.20 & 22.55 & 16.87 & 14.36 \\
\quad AvgLR Replay & 30.23 & 24.41 & 18.99 & 17.07 \\
\quad WSD                   & 29.18 & 22.47 & 16.62 & 14.31 \\
\quad CLR                   & 30.79 & 26.15 & 19.19 & 18.62 \\
\quad AutoLRS              & 30.07 & 23.63 & 18.95 & 16.46 \\
\quad MECHANIC              & 31.04 & 25.52 & 18.88 & 16.68 \\
\rowcolor{gray!10} \quad \textbf{SOLAR (online)} & \textbf{28.63} & \textbf{21.99} & \textbf{16.35} & \textbf{13.79} \\
\rowcolor{gray!10} \quad \textbf{SOLAR (frozen)} & -- & \textbf{21.87} & \textbf{16.22} & \textbf{13.54} \\
\bottomrule
\end{tabular}

\end{table}

Both SOLAR modes outperform every non-SOLAR alternative wherever they are evaluated. At 1B, SOLAR-online changes AdamW+Cosine from 16.52 to 14.83 and Muon+Cosine from 14.36 to 13.79. At the larger target scales, SOLAR-frozen further lowers five of the six seed-52 online results by reusing the source-acquired controller. Blockwise LR does not consistently improve the tuned base, while AutoLRS and MECHANIC trail the static schedules in most settings. Schedule-Free AdamW~\citep{defazio2024roadscheduled} improves slightly over Cosine at all four scales and is the strongest non-SOLAR alternative at 350M, yet remains above SOLAR. The gains persist across the repeated runs summarized in Appendix~\ref{app:seed_results}.

\subsubsection{MoE Pretraining under Non-Stationary Dynamics}

MoE training introduces routing-induced stochasticity and sparse, uneven gradient updates, which often require conservative optimization settings~\citep{fedus2022switchtransformersscalingtrillion, zoph2022stmoedesigningstabletransferable}. We evaluate SOLAR on Qwen2-MoE 1B~\citep{qwen2technicalreport} pretrained on The Pile under the setup in Appendix~\ref{app:moe_setup}. Figure~\ref{fig:moe_pretrain} shows a final PPL improvement from 9.61 to 9.34. SOLAR uses a higher average LR while producing a lower, smoother gradient-norm trace than Cosine in this run. At the module level (Figure~\ref{fig:moe_micro_lr}), the effective LRs fluctuate around a decaying profile, with distinct actions across groups. Section~\ref{sec:mechanistic_analysis} examines these traces. Appendix~\ref{app:megatron_3b_moe} reports the separate 3B DeepSeek-V2-style MoE run in Megatron, where final PPL changes from 10.73 to 10.38.

\begin{figure*}[t]
    \centering
    \begin{subfigure}[t]{0.48\textwidth}
        \centering
        \includegraphics[width=\linewidth]{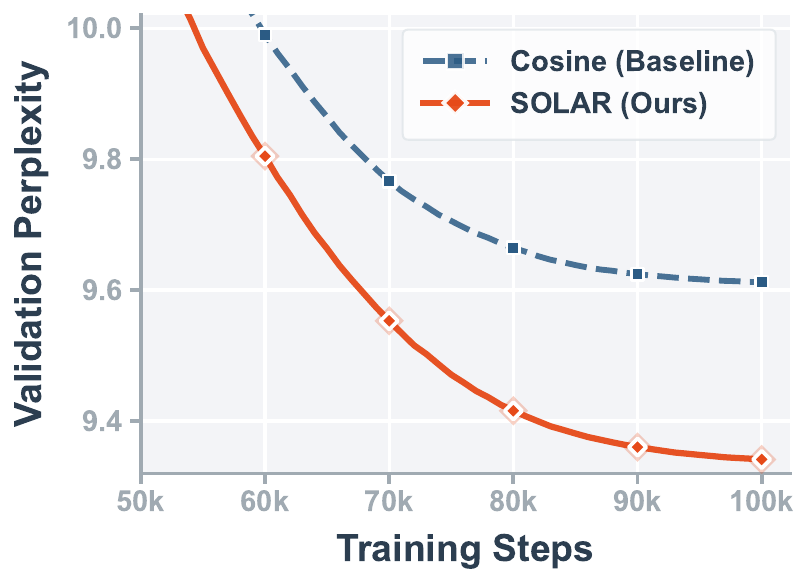}
        \caption{\small Validation PPL}
        \label{fig:moe_ppl}
    \end{subfigure}%
    \hfill
    \begin{subfigure}[t]{0.48\textwidth}
        \centering
        \includegraphics[width=\linewidth]{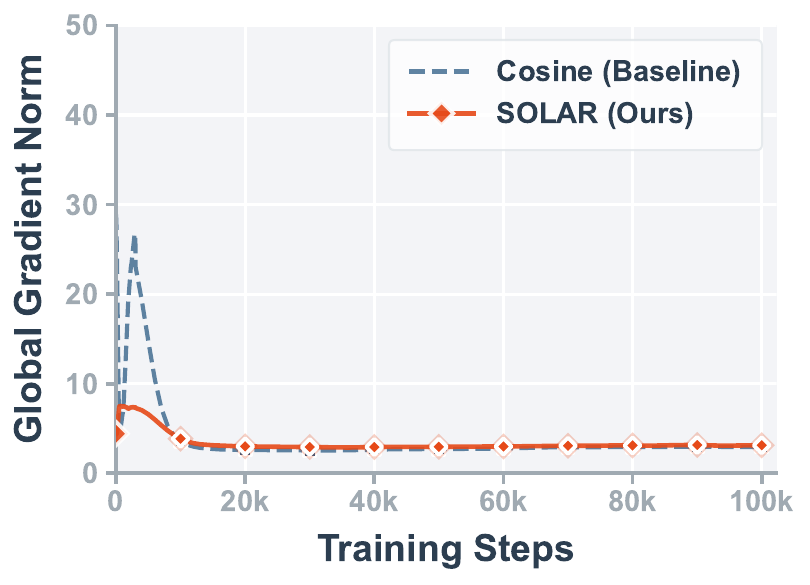}
        \caption{\small Global Grad Norm}
        \label{fig:moe_gradnorm}
    \end{subfigure}\\[6pt]
    \begin{subfigure}[t]{0.48\textwidth}
        \centering
        \includegraphics[width=\linewidth]{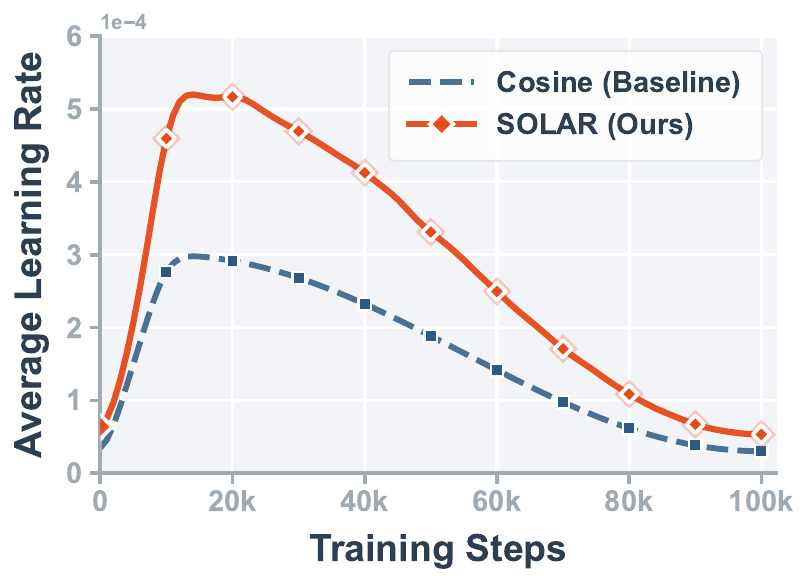}
        \caption{\small Macro: Average LR}
        \label{fig:moe_avg_lr}
    \end{subfigure}%
    \hfill
    \begin{subfigure}[t]{0.48\textwidth}
        \centering
        \includegraphics[width=\linewidth]{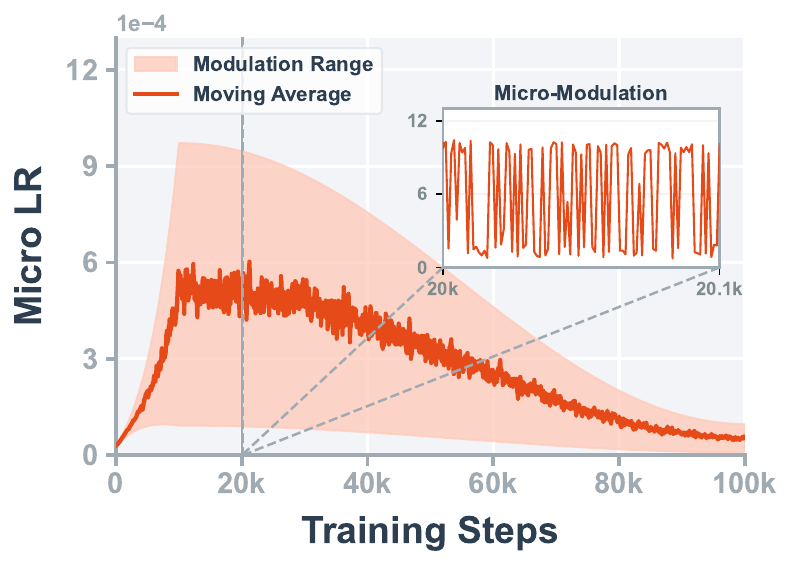}
        \caption{\small Micro: Attention LR}
        \label{fig:moe_micro_lr}
    \end{subfigure}
    \caption{\small Qwen2-MoE 1B pretraining on The Pile. SOLAR improves validation perplexity, suppresses early gradient-norm spikes, and maintains a higher average LR through group-wise micro-modulation. Panel~(d) shows one representative attention projection group; the solid line is a 200-step moving average and the shaded region visualizes the corresponding step-wise variation.}
    \label{fig:moe_pretrain}
\end{figure*}

\paragraph{Runtime and Circuit-Breaker usage.} No main-table or MoE run triggers rollback. On 1B AdamW, online SOLAR adds 1.23\% to full-run wall-clock time and 1.27\% to steady-state step time; the frozen mode adds 0.76\% and 0.81\%, respectively. Appendix~\ref{app:runtime_overhead} defines both measurements and reports the remaining settings.

\subsection{Frozen Residual-Policy Transfer Across Model Scales}
\label{sec:exp_transfer}

The residual parameterization expresses actions relative to the current base LR. We test whether this dimensionless policy can be acquired on a small proxy and reused at larger dense scales.

\textbf{Setup.} We train SOLAR on five 60M pretraining trajectories of 11K updates. The policy is then frozen and applied to 130M, 350M, and 1B models within the same optimizer family. Each target run combines the frozen policy with its base schedule, performs no PPO updates, and does not retune SOLAR-specific hyperparameters. All source-policy parameters, including the final learned shared $\log\sigma$, remain fixed, while group-wise actions are recomputed from the current target state at every update.

\textbf{Results.} SOLAR-frozen improves over the corresponding target-tuned Cosine and WSD schedules at every larger scale in Table~\ref{tab:main_results}. It yields the lower PPL in five of six matched seed-52 comparisons with SOLAR-online, while online acquisition yields the lower PPL under the corpus and optimizer shifts in Appendix~\ref{app:online_frozen_modes}. Across repeated 130M runs, SOLAR-online and SOLAR-frozen reach mean final PPLs of 22.86 and 22.76 with AdamW, and 22.05 and 22.09 with Muon, respectively. Frozen reuse improves from 25.04 PPL at random initialization to 22.41 after five full-length 60M source runs (Appendix~\ref{app:online_frozen_modes}). Appendix~\ref{app:transfer_robustness} tests the same policy under $0.5\times$, $1\times$, and $2\times$ target base LRs: it beats the matched Cosine run in all six settings and completes both $2\times$ runs where Cosine diverges. A separate $\mu$P width-transfer experiment removes target-scale base-LR search and retains the improvement, while a C4-to-Pile experiment tests simultaneous scale and corpus shift.

\subsection{Mechanistic Evidence}
\label{sec:mechanistic_analysis}

\begin{figure*}[t]
\centering
\begin{subfigure}[t]{0.48\textwidth}
  \centering
  \includegraphics[width=\linewidth]{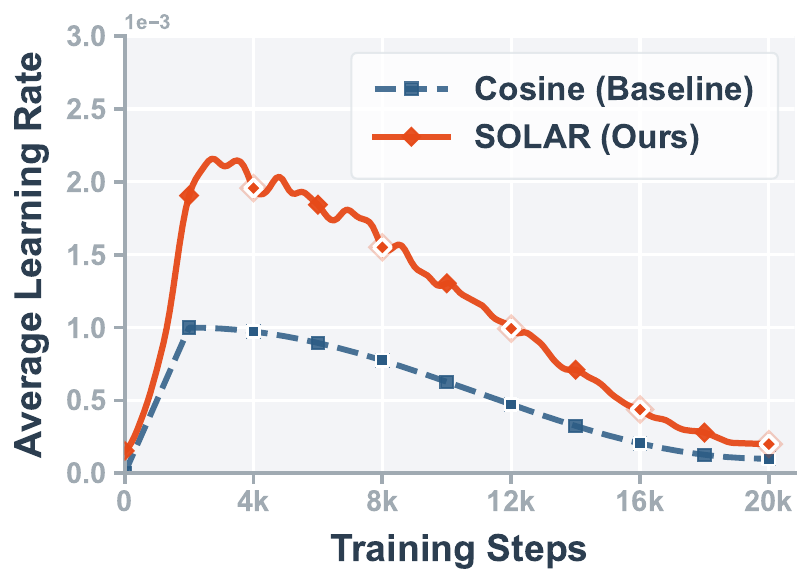}
  \caption{\small Average effective LR}
  \label{fig:mech_avg_lr}
\end{subfigure}%
\begin{subfigure}[t]{0.48\textwidth}
  \centering
  \includegraphics[width=\linewidth]{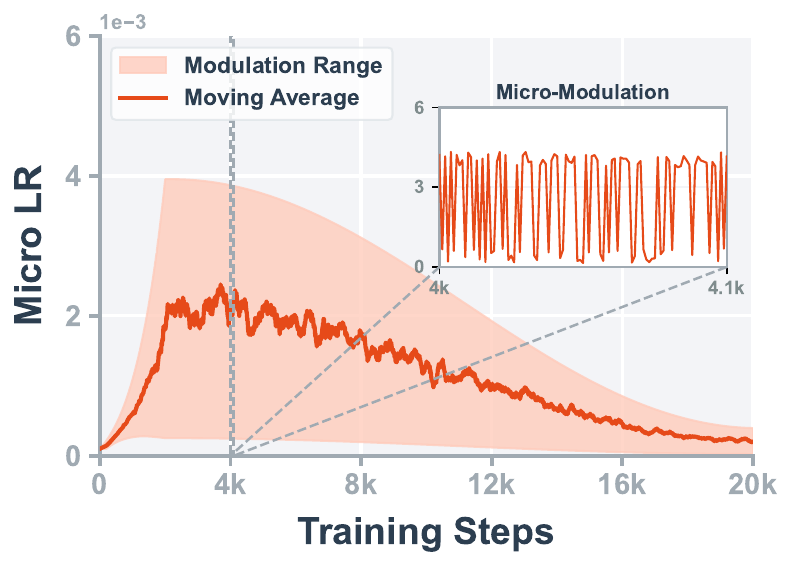}
  \caption{\small Single-group dynamics}
  \label{fig:mech_micro_lr}
\end{subfigure}
\\[6pt]
\begin{subfigure}[t]{0.48\textwidth}
  \centering
  \includegraphics[width=\linewidth]{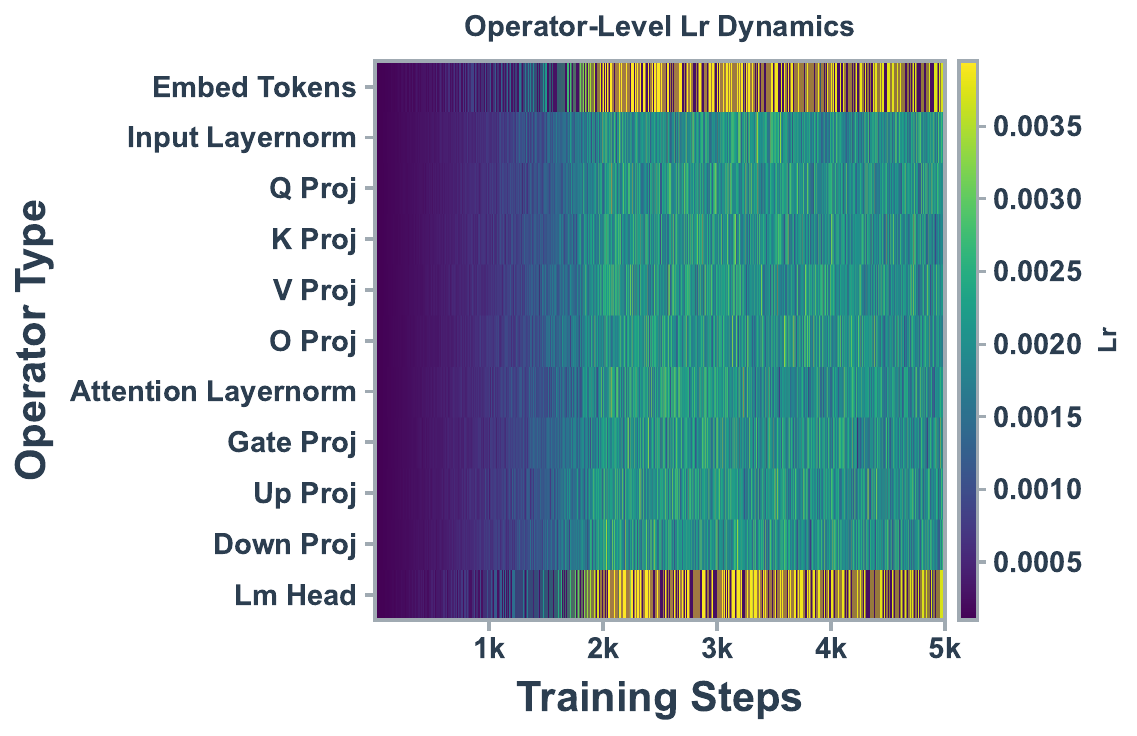}
  \caption{\small Module-wise allocation}
  \label{fig:mech_operator_heatmap}
\end{subfigure}%
\begin{subfigure}[t]{0.48\textwidth}
  \centering
  \includegraphics[width=\linewidth]{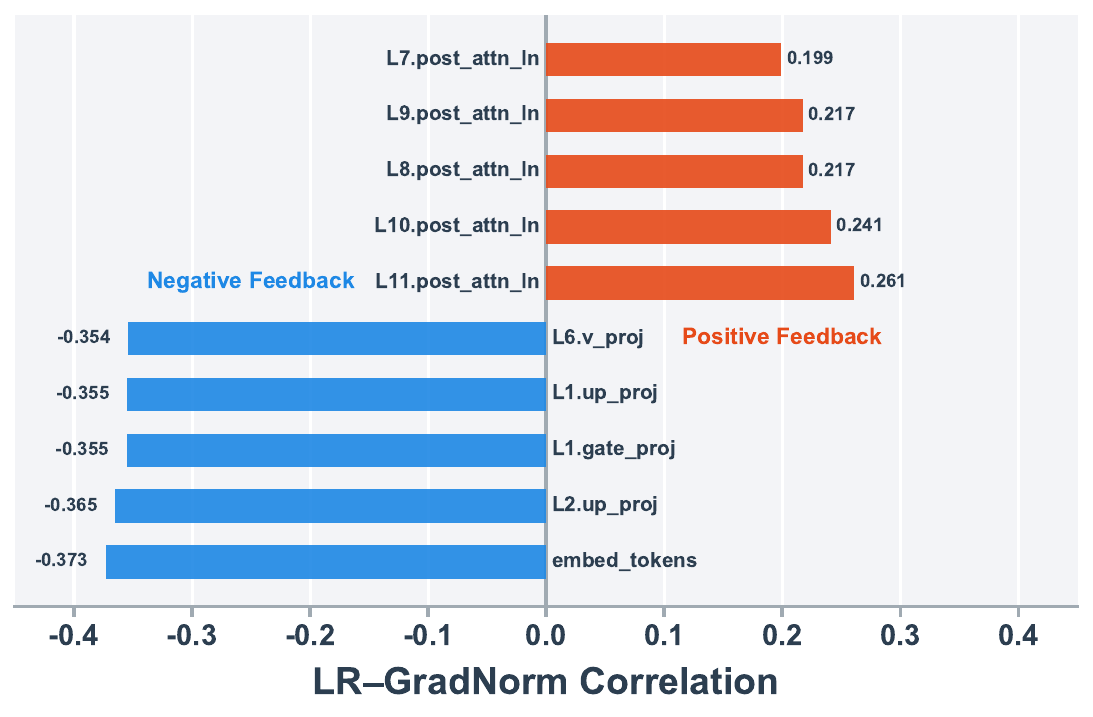}
  \caption{\small LR--GradNorm relation}
  \label{fig:mech_lr_gradnorm_corr}
\end{subfigure}

\caption{\small Mechanistic analysis of SOLAR on 130M Llama~2. (a)--(d): average effective LR, single-group LR dynamics, module-wise LR allocation, and per-module LR--GradNorm correlation, respectively. In (b), the solid line is a 200-step moving average and the shaded region visualizes the corresponding step-wise variation.}
\label{fig:solar_mechanistic_analysis}
\end{figure*}

Unless noted otherwise, we analyze the seed-52 130M Llama~2 run to characterize SOLAR's structured control. Figure~\ref{fig:solar_mechanistic_analysis} shows that SOLAR acts along three axes: temporal adaptation, module-wise redistribution, and module-specific feedback to gradient statistics.

\textbf{Control structure.} Matched controls identify which design choices separate improvement from degradation. A recursive global PPO controller reaches 27.09 final PPL. Re-anchoring and bounding the same global interface lowers PPL to 23.74; replacing the global action with group-wise control reaches 22.87 ($n=2$, seeds 42 and 52). Group-wise hypergradient control reaches 24.26, while an LLM adaptation of the layer-wise GANNO controller reaches 24.95. Untrained, near-deterministic, and state-blind SOLAR policies remain above 25.0. The anchored, state-conditioned global interface reaches 23.74, and group-wise control provides a further 0.87-PPL gain. Appendix~\ref{app:novelty_controls} gives the implementations, tuning protocol, and per-seed results.

\textbf{Temporal adaptation.} SOLAR maintains a higher average LR than Cosine, while AvgLR Replay ends at 25.61 PPL (Figure~\ref{fig:mech_avg_lr}). A normalized static group profile reaches 24.19 PPL, compared with 22.79 for SOLAR-online on the same seed. These controls show that neither a scalar temporal schedule nor a static group allocation reproduces SOLAR's online trajectory. Appendix~\ref{app:open_loop_replays} gives the protocols and results.

\textbf{Module-wise redistribution.} Figure~\ref{fig:mech_operator_heatmap} shows a stable module-level allocation pattern: vocabulary-facing components (token embeddings, LM head) receive larger multipliers, while internal attention and MLP projections are more constrained. This structure is consistent across training and suggests that SOLAR reallocates optimization capacity toward modules that appear more sensitive to learning-rate variation. Such structural allocation cannot be reproduced by a scalar global schedule.

\textbf{Module-specific feedback.} SOLAR learns differentiated LR--GradNorm relationships (Figure~\ref{fig:mech_lr_gradnorm_corr}). Dense transformation matrices exhibit negative correlation, while several post-attention LayerNorm parameters show positive correlation. The mapping therefore changes across modules rather than applying a uniform inverse function of gradient norm.

Appendix~\ref{app:additional_ablations} provides state, action, reward, Circuit-Breaker, and base-LR analyses. Appendix~\ref{app:novelty_controls} contains the matched learned-controller comparisons.

\section{Conclusion and Limitations}
\label{sec:conclusion}

We presented SOLAR, an online LR scheduler that uses a base schedule for the coarse LR profile while adapting to the realized optimization state. Its policy learns bounded, state-dependent group-wise corrections that re-anchor to the base at every step, guided by lightweight training features and a progress-aware reward. Across dense and MoE settings, SOLAR improves final PPL over tuned static schedules and automatic LR tuners under both AdamW and Muon, with repeated gains through 1B. On matched seeds, controls show that base anchoring and action bounds improve a global PPO controller from 27.09 to 23.74 PPL, while group-wise control reaches 22.87.

The controls also clarify how the design divides the scheduling problem. Re-anchoring preserves the base scheduler's warmup--decay profile, bounded residual actions limit the effect of individual policy decisions, and group-wise feedback allows the correction to differ across parameter tensors as the run evolves. This structure supports two complementary operating modes. SOLAR-online learns the controller within the current pretraining run, while SOLAR-frozen reuses an acquired policy without target-side PPO updates or SOLAR-specific retuning. A policy trained on a 60M proxy transfers to larger dense scales and beats matched Cosine runs across the tested fourfold base-LR range. The cross-corpus, reward, and $\mu$P results in the appendix extend the evaluation across changes in data, reward composition, and parameterization.

These results provide a practical route to adaptive LR control in LLM pretraining. The primary limitation of SOLAR is that its validation is still limited to dense models up to 1B parameters and a supplementary 3B MoE setting, so its behavior at larger scales remains untested. We leave this investigation to future work.

\bibliographystyle{iclr2027_conference}
\bibliography{reference}

\newpage
\appendix

\renewcommand{\appendixpagename}{Appendix}

\begin{center}
\LARGE \bfseries \appendixpagename
\end{center}

\section{SOLAR Implementation Details}
\label{app:rl_lrs_details}

In this section, we provide the complete implementation details of the SOLAR scheduler, including state construction, action parameterization, PPO training configurations, and the automated Circuit-Breaker mechanism.

\subsection{State Definition and Feature Extraction}
\label{app:state_impl}

This appendix expands the compact state definition in Section~\ref{subsec:state} by specifying normalization constants, EMA coefficients, window sizes, and missing-gradient handling.

\paragraph{State definition.}
For controlled parameter group $g$, the scheduler state is defined as the concatenation of global optimization features and local group-specific features:
\begin{equation*}
    s_{t,g} = \left[ s_t^{\mathrm{global}};\; s_{t,g}^{\mathrm{local}} \right].
\end{equation*}

\paragraph{Global features.}
The global state captures the overall training dynamics:
\begin{equation*}
    s_t^{\mathrm{global}} = \left[ \tau_t,\; \log L_t,\; \nu_t,\; \Delta \mathrm{EMA}_t \right].
\end{equation*}
In implementation, the normalized training progress is $\tau_t = t/T$, where $T$ is the total number of training steps. The loss term is implemented as:
\begin{equation*}
    \log L_t = \log(L_t + \epsilon),
\end{equation*}
where $\epsilon = 10^{-8}$ is the numerical stability constant (used uniformly across all log and EMA denominator computations in this appendix). The short-window loss fluctuation is computed from a sliding window of recent losses:
\begin{equation*}
    \nu_t = \frac{\operatorname{Std}(\mathcal H_t)}{L_t + \epsilon},
\end{equation*}
where $\mathcal H_t$ stores the most recent $w=20$ losses; when fewer than $20$ steps have elapsed, we use all available samples. To characterize short-term versus long-term optimization trends, we maintain two exponential moving averages (EMA) of the loss with $\beta_s=0.9$ and $\beta_\ell=0.99$, initialized as $\mathrm{EMA}^{\mathrm{short}}_0 = \mathrm{EMA}^{\mathrm{long}}_0 = L_0$:
\begin{align}
    \mathrm{EMA}^{\mathrm{short}}_t &= \beta_s \mathrm{EMA}^{\mathrm{short}}_{t-1} + (1-\beta_s)L_t, \\
    \mathrm{EMA}^{\mathrm{long}}_t &= \beta_\ell \mathrm{EMA}^{\mathrm{long}}_{t-1} + (1-\beta_\ell)L_t.
\end{align}
The trend difference is then defined as:
\begin{equation*}
    \Delta \mathrm{EMA}_t = \frac{\mathrm{EMA}^{\mathrm{short}}_t - \mathrm{EMA}^{\mathrm{long}}_t}{\mathrm{EMA}^{\mathrm{long}}_t + \epsilon}.
\end{equation*}

\paragraph{Local features.}
The local state captures tensor-specific statistics:
\begin{equation*}
    s_{t,g}^{\mathrm{local}} = \left[ \log \eta^{\mathrm{base}}_{t,g},\; \log \|\nabla_{t,g}\|,\; a_{t-1,g},\; d_g,\; \log \|w_{t,g}\|,\; \Delta \log \|\nabla_{t,g}\| \right].
\end{equation*}
The base learning-rate context is $\log \eta^{\mathrm{base}}_{t,g} = \log(\eta^{\mathrm{base}}_{t,g} + \epsilon)$ and the parameter-scale feature is $\log \|w_{t,g}\| = \log(\|w_{t,g}\|_2 + \epsilon)$, both using the same $\epsilon = 10^{-8}$. The gradient-magnitude feature and its recent change are computed respectively as:
\begin{align}
    \log \|\nabla_{t,g}\| &= \log(\|\nabla_{t,g}\|_2 + \epsilon), \\
    \Delta \log \|\nabla_{t,g}\| &= \log(\|\nabla_{t,g}\|_2 + \epsilon) - \log(\|\nabla_{t-1,g}\|_2 + \epsilon).
\end{align}
The previous action $a_{t-1,g}$ is the action applied to controlled group $g$ at the previous step. The depth feature $d_g$ records the normalized network depth of the module containing controlled group $g$. If a controlled parameter group has no gradient at step $t$, we set gradient-dependent local entries ($\log\|\nabla_{t,g}\|$, $\Delta\log\|\nabla_{t,g}\|$) to zero and keep non-gradient entries ($\log\eta^{\mathrm{base}}_{t,g}$, $a_{t-1,g}$, $d_g$, $\log\|w_{t,g}\|$) unchanged when available.
Each state feature is normalized with EMA running statistics using decay 0.99 before it is passed to the policy.

\subsection{Action Space, Architecture, and Distributed Execution}
\label{app:action_ppo_impl}

\paragraph{Action Parameterization.}
SOLAR controls hyperparameters at the fine-grained level of controlled parameter groups. Each controlled group corresponds to one trainable parameter tensor in the dense implementation; the MoE implementations use the module partition defined by their respective codebases. For every group, the actor samples a pre-Tanh variable, forms a squashed action, and clips only the value sent to the optimizer:
\begin{equation}
\begin{aligned}
u_{t,g} &\sim \mathcal N(\mu_{t,g},\sigma^2), \\
\widetilde a_{t,g} &= \tanh(u_{t,g}), \\
a_{t,g} &= \operatorname{clip}(\widetilde a_{t,g},-1+10^{-4},1-10^{-4}).
\end{aligned}
\label{eq:action_variables}
\end{equation}
The PPO log-probability is evaluated from the stored $u_{t,g}$ with the Tanh change-of-variables correction, before the numerical clipping that produces $a_{t,g}$.

The scheduler applies multiplicative residuals to the underlying base scheduler. The executed learning rate for controlled parameter group $g$ at step $t$ is:
\begin{equation}
    \eta_{t,g} = \eta_{t,g}^{\mathrm{base}} \cdot \exp\left(\alpha_t a_{t,g}\right),
    \label{eq:exec_lr_impl}
\end{equation}
where $\eta_{t,g}^{\mathrm{base}}$ is the base learning rate and $\alpha_t$ is the action scale. To ensure stability during the initial phase, we linearly warm up the action scale: $\alpha_t = \alpha \cdot (t / T_{\mathrm{warmup}})$ for $t < T_{\mathrm{warmup}}$, and $\alpha_t = \alpha$ afterward, where $T_{\mathrm{warmup}}=0.1T$ across all experiments. We set the maximum action scale to $\alpha = 1.3$.

\paragraph{Architecture and PPO Setup.}
We instantiate SOLAR with a lightweight actor-critic network. Given the state representation \(s_{t,g}\) for controlled parameter group \(g\), a shared MLP encoder first produces a hidden representation:
\[
h_{t,g} = f_{\mathrm{enc}}(s_{t,g}),
\]
where \(f_{\mathrm{enc}}\) is a two-layer MLP with input dimension $10$ (matching the state dimension), hidden dimension $256$, and Tanh activations; no layer normalization is used. This representation is then passed to separate actor and critic heads, which are implemented as linear projections:
\[
\mu_{t,g} = W_\mu h_{t,g} + b_\mu, \qquad V_\phi(s_{t,g}) = W_v h_{t,g} + b_v.
\]
The actor policy \(\pi_\theta(\widetilde a_{t,g}\mid s_{t,g})\) is a one-dimensional Tanh-Normal distribution with mean \(\mu_{t,g}\) and a globally learnable scalar log-standard-deviation parameter \(\log \sigma\) shared across all controlled parameter groups.

The squashed action distribution factorizes across controlled groups,
\begin{equation}
\pi_\theta(\widetilde a_t \mid s_t)
=
\prod_{g=1}^{G}\pi_\theta(\widetilde a_{t,g}\mid s_{t,g}),
\label{eq:factorized_policy}
\end{equation}
where all factors share the actor parameters \(\theta\). For the stored pre-Tanh sample $u_{t,g}$, define
\begin{equation}
\ell_{t,g}(\theta)
=
\log \mathcal N\!\left(u_{t,g};\mu_\theta(s_{t,g}),\sigma^2\right)
-\log\!\left(1-\tanh^2(u_{t,g})\right),
\label{eq:group_logprob}
\end{equation}
which is the Tanh-Normal log-probability evaluated from the stored $u_{t,g}$. All action factors share \(\theta\) and act on the same LLM transition. Policy learning uses a parameter-shared independent update. Let \(\theta_{\mathrm{old}}\) denote the behavior-policy parameters used to collect the rollout. For group \(g\), the other action factors remain at their behavior-policy samples while its local objective is evaluated under \(\theta\):
\begin{equation}
J_g^{\mathrm{ind}}(\theta;\theta_{\mathrm{old}})
=
\mathbb E_{\substack{
\widetilde a_{\tau,g}\sim\pi_\theta(\cdot\mid s_{\tau,g})\\
\widetilde a_{\tau,-g}\sim\pi_{\theta_{\mathrm{old}}}(\cdot\mid s_{\tau,-g})
}}
\left[
\sum_{\tau=t}^{T-1}\gamma^{\tau-t}r^{\mathrm{final}}_{\tau+1,g}
\right].
\label{eq:independent_group_objective}
\end{equation}
Using group-wise GAE, the independent policy update uses the gradient estimate
\begin{equation}
\widehat g_g
:=
\widehat{\mathbb E}_t\!\left[
\hat A_{t,g}
\left.\nabla_\theta\ell_{t,g}(\theta)\right|_{\theta=\theta_{\mathrm{old}}}
\right].
\label{eq:independent_group_gradient}
\end{equation}
PPO applies likelihood-ratio clipping to each group objective. The group-specific ratio is
\begin{equation}
\rho_{t,g}(\theta)
=
\exp\!\left(
\ell_{t,g}(\theta)-\ell_{t,g}(\theta_{\mathrm{old}})
\right).
\label{eq:group_policy_ratio}
\end{equation}
The per-group log-probabilities are not summed when computing \(\rho_{t,g}\). Averaging the group surrogates gives
\begin{equation}
\mathcal L_{\mathrm{PPO}}(\theta)
=
\mathbb{E}_t\!\left[
\frac{1}{G}\sum_{g=1}^{G}
\min\!\left(
\rho_{t,g}(\theta)\hat A_{t,g},
\operatorname{clip}\!\left(\rho_{t,g}(\theta),1-\epsilon_{\mathrm{clip}},1+\epsilon_{\mathrm{clip}}\right)\hat A_{t,g}
\right)
\right],
\label{eq:ppo_obj}
\end{equation}
where \(\hat A_{t,g}\) is the generalized advantage estimate (GAE)~\citep{schulman2018highdimensionalcontinuouscontrolusing} for group \(g\), and \(\epsilon_{\mathrm{clip}}\) is the clipping threshold. The minimum and clipping operations are applied to each \((t,g)\) entry before the shared actor receives their average. The value and entropy terms use the same group-wise reduction:
\begin{align}
\mathcal L_V(\phi)
&=
\mathbb E_t\!\left[
\frac{1}{G}\sum_{g=1}^{G}
\left(V_\phi(s_{t,g})-\hat R_{t,g}\right)^2
\right],
\label{eq:group_value_loss}\\
\mathcal L_H(\theta)
&=
\mathbb E_t\!\left[
\frac{1}{G}\sum_{g=1}^{G}
\mathcal H\!\left[\pi_\theta(\cdot\mid s_{t,g})\right]
\right].
\label{eq:group_entropy}
\end{align}
The actor--critic minimizes
\begin{equation}
\mathcal J(\theta,\phi)
=
-\mathcal L_{\mathrm{PPO}}(\theta)
+\lambda_v\mathcal L_V(\phi)
-\lambda_e\mathcal L_H(\theta),
\label{eq:actor_critic_objective}
\end{equation}
where \(\hat R_{t,g}\) is the target return, \(\lambda_v\) is the value-loss coefficient, and \(\lambda_e\) is the entropy coefficient.

In our implementation, we optimize \(\mathcal{J}(\theta, \phi)\) using the Adam optimizer with a learning rate of \(3 \times 10^{-4}\). Standard PPO hyperparameters include a discount factor \(\gamma = 0.99\), a clipping threshold \(\epsilon_{\mathrm{clip}} = 0.2\), a value loss coefficient \(\lambda_v = 0.5\), and an entropy coefficient \(\lambda_e = 0.05\). During normal training, the RL agent collects trajectories and performs an update every 50 steps, with \(K=4\) optimization epochs per update.

For a rollout of \(T_r\) decision steps, states have shape \([T_r,G,d_s]\); pre-Tanh samples, executed actions, log-probabilities, ratios, returns, values, and advantages have shape \([T_r,G]\). GAE is computed independently along the temporal axis for each controlled group with \(\lambda_{\mathrm{GAE}}=0.95\). Advantages are normalized once over the complete \([T_r,G]\) block as \(\texttt{adv} \leftarrow (\texttt{adv} - \texttt{mean}(\texttt{adv})) / (\texttt{std}(\texttt{adv}) + 10^{-7})\) and reused across PPO epochs. Reward normalization and value-loss clipping are not used.

The actor-critic network weights are initialized as: all feature and hidden linear layers use orthogonal initialization with gain \(\sqrt{2}\), biases set to \(0\); the actor output layer uses constant initialization of \(0.01\) on weights with zero bias, so that the initial mean \(\mu_{t,g}\approx 0\); the critic output layer uses orthogonal initialization with gain \(\sqrt{2}\) and zero bias; the log-standard-deviation is initialized to \(0\) (yielding \(\sigma=1.0\)) and is learned. Because the scheduler only predicts low-dimensional multiplicative LR corrections while leaving the base optimizer's internal mechanics unchanged, SOLAR remains highly lightweight and seamlessly integrates into standard pretraining pipelines.

\paragraph{Base Scheduler and Transfer Configuration.}
For the reported dense experiments, the base scheduler uses the Cosine peak LR listed for the corresponding model scale and optimizer in Appendix~\ref{app:tuning_protocol}. The Qwen2-MoE experiment also uses a Cosine base, while the supplementary 3B MoE experiment uses WSD as specified in Appendix~\ref{app:megatron_3b_moe}. In frozen residual-policy transfer, the target combines its corresponding base schedule with the source policy. The policy performs no target-scale PPO updates; the source policy's final learned parameters, including the shared $\log\sigma$, are transferred and kept fixed together with $\alpha$, action warmup, reward coefficients, and PPO settings.

\paragraph{Distributed Execution.}
To minimize synchronization overhead during large-scale pretraining, the RL agent operates exclusively on the rank-0 worker. The rank-0 process evaluates the policy, samples actions, and updates the PPO agent. The sampled actions are then broadcast to all other workers to ensure identical optimizer states across the distributed cluster. In the Distributed Data Parallel (DDP) setting, gradient norms are computed locally on each rank from the already-synchronized gradients (all-reduced during \texttt{backward()}); per-parameter norms are then broadcast from rank-0 rather than all-reduced, since the gradient values are already global and computing norms is a secondary reduction. The full procedure is summarized in Algorithm~\ref{alg:rllrs_step}.

\subsection{Reward Design and Circuit-Breaker Mechanism}
\label{app:reward_cb_impl}

Let \(L_t\) and \(\nabla_t\) denote the loss and synchronized gradient from the forward/backward \emph{before} the optimizer update at step \(t\). The scheduler observes state \(s_t\), samples action \(a_t\), and the optimizer updates parameters. After the update, the next forward/backward produces \(L_{t+1}\) and post-update gradient \(\nabla_{t+1}\). The reward credited to \(a_t\) is then computed as follows. Since the scheduler outputs controlled-parameter-group-wise learning-rate adjustments, each controlled parameter group receives a reward that combines global progress feedback with a controlled-group-wise stability shaping term:
\begin{equation}
\label{eq:reward_decomp}
r_{t+1,g}
=
r_{t+1}^{\mathrm{perf}}
+
r_{t+1}^{\mathrm{trend}}
-
p_{t+1,g}^{\mathrm{stab}}.
\end{equation}

\paragraph{Performance term.}
We first measure the relative improvement over the previous step:
\begin{equation}
\label{eq:chi_perf}
\chi_{t+1}
=
\frac{L_t}{L_{t+1} + 10^{-10}}.
\end{equation}
The performance term is defined as:
\begin{equation}
\label{eq:r_perf}
r_{t+1}^{\mathrm{perf}}
=
20 \cdot \log(\chi_{t+1}).
\end{equation}

\paragraph{Trend term.}
We incorporate a smoother long-term signal based on the long EMA of the loss:
\begin{equation}
\label{eq:r_trend}
r_{t+1}^{\mathrm{trend}}
=
2 \cdot
\frac{\mathrm{EMA}^{\mathrm{long}}_{t+1} - L_{t+1}}
{\mathrm{EMA}^{\mathrm{long}}_{t+1} + 10^{-8}}.
\end{equation}

\paragraph{Controlled-group-wise stability shaping term.}
We compute stability feedback at the same granularity as the scheduling actions. For each controlled parameter group \(g\), let \(\|\nabla_{t+1,g}\|_2\) denote its gradient norm from the backward \emph{after} the optimizer update, and let \(m_{t+1,g}\) be its exponential moving average (EMA):
\begin{equation*}
m_{t+1,g} = \beta m_{t,g} + (1-\beta)\|\nabla_{t+1,g}\|_2,
\end{equation*}
where \(\beta=0.99\) is the smoothing coefficient. The EMA is initialized from the first observed gradient norm, \(m_{0,g}=\|\nabla_{0,g}\|_2\). We define the relative gradient-growth ratio as
\begin{equation}
\label{eq:q_stab}
q_{t+1,g} = \frac{\|\nabla_{t+1,g}\|_2}{m_{t+1,g} + \epsilon},
\end{equation}
where \(\epsilon = 10^{-8}\) is the numerical stability constant.
The controlled-group-wise stability shaping term is
\begin{equation}
\label{eq:p_stab}
p_{t+1,g}^{\mathrm{stab}} =
\begin{cases}
q_{t+1,g}-1, & q_{t+1,g} \le \tau_{\mathrm{sev}}, \\[4pt]
q_{t+1,g}-1 + \delta_{\mathrm{sev}}, & q_{t+1,g} > \tau_{\mathrm{sev}},
\end{cases}
\end{equation}
where \(\tau_{\mathrm{sev}}=3.0\) and \(\delta_{\mathrm{sev}}=20\).

This signed term compares each group's post-update gradient norm with its own recent trend. Since it is subtracted from the reward, \(q_{t+1,g}<1\) provides positive shaping, whereas \(q_{t+1,g}>1\) reduces the reward; the severe offset applies when \(q_{t+1,g}>\tau_{\mathrm{sev}}\). Because the signal is computed after the update and indexed by controlled group, it provides group-resolved feedback aligned with the group-wise actions. The two progress terms remain shared across groups and tied to the global language-modeling objective. Since the current gradient norm also enters \(m_{t+1,g}\), \(0\le q_{t+1,g}\le 1/(1-\beta)=100\) and \(-1\le p_{t+1,g}^{\mathrm{stab}}\le119\); no additional clipping is used.

\paragraph{Circuit-Breaker and Restart Procedure.}
To prevent catastrophic divergence during online exploration and ensure training stability, we implement an automated Circuit-Breaker mechanism. Unlike the controlled-group-wise stability shaping term, which provides dense feedback during normal optimization, the Circuit-Breaker is a global emergency mechanism triggered by severe loss spikes. At each step, we compute the loss spike ratio:
\begin{equation}
\label{eq:loss_spike_ratio}
\kappa_{t+1}
=
\frac{L_{t+1}}
{L^{\mathrm{ema}}_{t+1} + 10^{-8}}.
\end{equation}
If \(\kappa_{t+1} > \kappa_L\), the Circuit-Breaker is triggered, where \(\kappa_L=1.5\). The optimizer wrapper executes the following procedure:
\begin{enumerate}
    \item \textbf{Punishment:} Applies the global Circuit-Breaker penalty
    \(r^{\mathrm{final}}_{t+1,g}=r_{t+1,g}-\lambda_{\mathrm{cb}}\) for all controlled parameter groups \(g\),
    with \(\lambda_{\mathrm{cb}}=100\).
    \item \textbf{Forced Update:} Rank 0 immediately performs a PPO update using the transitions accumulated so far, ensuring the agent promptly learns from this catastrophic failure event.
    \item \textbf{Abort Signal:} Sets an \texttt{abort\_training=True} flag and broadcasts it to all workers.
\end{enumerate}
The optimizer wrapper does not restore model weights. It returns an abort signal to the outer training loop, which truncates the corrupted trajectory and restores the latest safe model and optimizer state. The forced PPO update is retained rather than replaced by the checkpointed PPO state.

\paragraph{Hyperparameter settings.}
The numerical constants in \eqref{eq:reward_decomp}--\eqref{eq:loss_spike_ratio} are fixed design choices with distinct meanings. The coefficient \(20\) in \(r_{t+1}^{\mathrm{perf}}\) scales the stepwise improvement signal, and the coefficient \(2\) in \(r_{t+1}^{\mathrm{trend}}\) scales the long-term trend signal. The stability shaping term uses \(\tau_{\mathrm{sev}}=3.0\) and adds \(\delta_{\mathrm{sev}}=20\) when gradient growth crosses this threshold. The Circuit-Breaker uses \(\kappa_L=1.5\) as the loss-spike trigger and \(\lambda_{\mathrm{cb}}=100\) as the emergency penalty. In our main training runs, the Circuit-Breaker is not triggered, so \(\kappa_L\) and \(\lambda_{\mathrm{cb}}\) do not affect the nominal optimization trajectory. We selected this configuration once on a small proxy setting and kept it fixed across all experiments.

\subsubsection{Controller Coefficient Census and Reward Leave-One-Out Analysis}
\label{app:reward_analysis}

The calibration covers four reward coefficients and the short-loss EMA decay used in the state. Table~\ref{tab:reward_census} lists these coefficients together with the fixed state-normalization decay. Sweeping one coefficient at a time produces final PPL between 22.80 and 22.85, compared with 22.79 at the default point. The largest deviation, 0.06 PPL, is below the observed 130M SOLAR seed standard deviation of 0.08.

\begin{table}[htbp]
\centering
\footnotesize
\caption{\small Controller-coefficient census on 130M/C4/AdamW.}
\label{tab:reward_census}
\begin{tabularx}{\linewidth}{l c c X}
\toprule
Quantity & Default & Status & Values evaluated \\
\midrule
Short-term progress scale & 20 & tunable & $\{10,20,40\}$ and removal \\
EMA trend scale & 2 & tunable & $\{1,2,4\}$ and removal \\
Severe stability offset & 20 & tunable & $\{10,20,40\}$ and removal \\
Severe stability threshold & 3.0 & tunable & $\{2.5,3.0,4.0\}$ \\
State short-loss EMA decay $\beta_s$ & 0.90 & tunable & $\{0.90,0.95\}$ \\
State-normalization EMA decay & 0.99 & fixed & fixed across experiments \\
\bottomrule
\end{tabularx}
\end{table}

The coefficient sweep tests local calibration, while the leave-one-out experiment measures the contribution of each reward component. Removing immediate progress, trend, and stability increases final PPL by 1.94, 0.90, and 0.30, respectively. Removing stability also causes the only Circuit-Breaker activation observed under nominal hyperparameters.

\begin{table}[htbp]
\centering
\caption{\small Reward leave-one-out on 130M/C4/AdamW using seeds 42 and 52.}
\label{tab:reward_loo}
\begin{tabular}{lccccl}
\toprule
Variant & Seed 52 & Seed 42 & Mean PPL & $\Delta$ vs. full & CB activations \\
\midrule
Full reward & 22.79 & 22.95 & 22.87 & 0.00 & 0/2 \\
$-$ immediate progress & 24.91 & 24.71 & 24.81 & +1.94 & 0/2 \\
$-$ EMA trend & 23.71 & 23.83 & 23.77 & +0.90 & 0/2 \\
$-$ stability shaping term & 23.24 & 23.09 & 23.17 & +0.30 & 1/2 \\
\bottomrule
\end{tabular}
\end{table}

The activation occurs in the no-stability seed-42 run at update 12,340. The Circuit-Breaker restores the latest safe state, after which the run completes. Removing stability raises mean PPL by 0.30 and produces the only recovery event observed under the nominal settings.

\begin{algorithm}[H]
\caption{\small SOLAR optimizer step (PyTorch-style pseudocode)}
\label{alg:rllrs_step}
\SetAlgoLined
\KwIn{Model parameters \(\{w_g\}_{g=1}^G\), current loss \(L_t\), base scheduler \(f_{\mathrm{base}}\), policy \(\pi_\theta\)}

\tcp{From the forward/backward on the current batch}
\(L_t \leftarrow \textsc{BroadcastFromRank0}(L_t)\)\;
Update loss history and short/long EMAs\;
Compute global state \(s_t^{\mathrm{global}}\)\;

\For{\(g = 1\) \KwTo \(G\)}{
    Compute local features from \(\nabla_{t,g}\), \(w_{t,g}\), \(\eta^{\mathrm{base}}_{t,g}\), \(a_{t-1,g}\), and \(d_g\)\;
    Form state \(s_{t,g}\)\;
}

\eIf{\(\mathrm{rank} == 0\)}{
    \(\{u_{t,g},a_{t,g},\ell^{\mathrm{old}}_{t,g}\}_{g=1}^G \leftarrow \pi_\theta.\textsc{Act}(\{s_{t,g}\}_{g=1}^G)\)\;
}{
    Allocate placeholder actions\;
}

\(\{a_{t,g}\}_{g=1}^G \leftarrow \textsc{BroadcastFromRank0}(\{a_{t,g}\}_{g=1}^G)\)\;
Obtain \(\{\eta_{t,g}^{\mathrm{base}}\}_{g=1}^G\) from \(f_{\mathrm{base}}\)\;

\For{\(g = 1\) \KwTo \(G\)}{
    Set \(\eta_{t,g} \leftarrow \eta_{t,g}^{\mathrm{base}} \exp(\alpha_t a_{t,g})\)\;
}

\textsc{Optimizer.Step}()\;
\tcp{The next standard training iteration yields \(L_{t+1}, \nabla_{t+1}\); no additional forward/backward is run}
Construct next state \(s_{t+1}\) from \(L_{t+1}\) and \(\nabla_{t+1}\)\;
Compute controlled-group-wise rewards \(\{r_{t+1,g}\}_{g=1}^{G}\) using \(L_t\), \(L_{t+1}\), and \(\nabla_{t+1}\)\;
Compute \(\psi_{t+1}\) using Eq.~\eqref{eq:psi_indicator} and set \(r^{\mathrm{final}}_{t+1,g}\leftarrow r_{t+1,g}-\lambda_{\mathrm{cb}}\psi_{t+1}\) for every group \(g\)\;

\If{\(\mathrm{rank} == 0\)}{
    Store \((s_t,u_t,a_t,\ell^{\mathrm{old}}_t,r^{\mathrm{final}}_{t+1},s_{t+1})\), retaining the group axis, in the PPO buffer\;
    \If{update interval reached \textbf{or} Circuit-Breaker triggered}{
        \(\pi_\theta.\textsc{Update}()\)\;
    }
}

\If{Circuit-Breaker triggered (\(\psi_{t+1} = 1\))}{
    \textcolor{red!80!black}{\tcp{Abort Training}}
    Trigger abort signal and broadcast to all workers\;
}
\end{algorithm}

\section{Experimental Details}
\label{app:Experimental_Details}

\subsection{Dense Model Setup}
\label{app:dense_setup}

We follow the standardized experimental setup introduced by \citet{zhao2024galorememoryefficientllmtraining}, with our own scheduler implementations integrated into the same training framework.

For the Llama~2~\citep{touvron2023llama2openfoundation} models, we use C4~\citep{C4} as the pretraining corpus. We fix the batch size to 512 and the maximum sequence length to 256 for all Llama~2 experiments.

We train the following model scales:
\begin{itemize}
    \item 60M: 11K training steps, corresponding to 1.4B tokens;
    \item 130M: 20K training steps, corresponding to 2.6B tokens;
    \item 350M: 60K training steps, corresponding to 7.8B tokens;
    \item 1B: 100K training steps, corresponding to 13.1B tokens.
\end{itemize}

Table~\ref{tab:llama_arch} summarizes the architectural configurations of all Llama~2 dense models used in our experiments. 
\begin{table}[H]
\centering
\caption{\small Architecture of Llama~2 dense models.}
\label{tab:llama_arch}
\renewcommand{\arraystretch}{1.15}
\setlength{\tabcolsep}{8pt}
\begin{tabular*}{\textwidth}{@{\extracolsep{\fill}}lcccc}
\toprule
& \textbf{60M} & \textbf{130M} & \textbf{350M} & \textbf{1B} \\
\midrule
\(d_{\text{model}}\) & 512 & 768 & 1024 & 2048 \\
\(n_{\text{layers}}\) & 8 & 12 & 24 & 24 \\
\(n_{\text{heads}}\) & 8 & 12 & 16 & 32 \\
\(d_{\text{ff}}\) & 1376 & 2048 & 2736 & 5461 \\
Vocab & 32000 & 32000 & 32000 & 32000 \\
\bottomrule
\end{tabular*}
\end{table}

For hyperparameter tuning, we sweep the peak learning rate over the following search grid:
\begin{equation*}
\{10^{-4},\, 3\times10^{-4},\, 5\times10^{-4},\, 10^{-3},\, 3\times10^{-3},\, 5\times10^{-3},\, 10^{-2}\}.
\end{equation*}
Alongside the learning rate, we also perform a grid search for the weight decay parameter over:
\begin{equation*}
\{10^{-2},\, 5\times 10^{-2},\, 10^{-1},\, 2\times 10^{-1}\}.
\end{equation*}

\subsection{Hyperparameter Selection and Final-PPL Reporting}
\label{app:tuning_protocol}

We use seed 42 to select scheduler configurations and keep the selected values fixed thereafter. Table~\ref{tab:main_results} reports validation PPL at the final update of the seed-52 runs. We evaluate validation PPL every 1K updates, but earlier checkpoints are not used for reporting. Training ends after 11K, 20K, 60K, and 100K updates for the 60M, 130M, 350M, and 1B models. Table~\ref{tab:seed_results} lists the repeated runs separately, including the seed-42 tuning runs.

All dense methods share sequence length 256, global batch size 512, gradient clipping at 1.0, and bf16 arithmetic. AdamW uses $\beta=(0.9,0.999)$ and $\epsilon=10^{-8}$; Muon uses momentum 0.95, with $\beta=(0.9,0.95)$ for its AdamW-handled parameters. The common base search covers peak LR $\{10^{-4},3\!\times\!10^{-4},5\!\times\!10^{-4},10^{-3},3\!\times\!10^{-3},5\!\times\!10^{-3},10^{-2}\}$ and weight decay $\{0.01,0.05,0.1,0.2\}$. Table~\ref{tab:scale_base_lrs} gives the selected base settings.

\begin{table}[t]
\centering
\caption{\small Selected peak LR for each dense-model scale. All settings use weight decay 0.01 and 10\% warmup. SOLAR inherits the corresponding Cosine base and does not run a separate base search.}
\label{tab:scale_base_lrs}
\begin{tabular}{lcccc}
\toprule
Optimizer & 60M & 130M & 350M & 1B \\
\midrule
AdamW & $3\!\times\!10^{-3}$ & $1\!\times\!10^{-3}$ & $1\!\times\!10^{-3}$ & $5\!\times\!10^{-4}$ \\
Muon  & $5\!\times\!10^{-3}$ & $3\!\times\!10^{-3}$ & $3\!\times\!10^{-3}$ & $1\!\times\!10^{-3}$ \\
\bottomrule
\end{tabular}
\end{table}

\begin{table}[t]
\centering
\footnotesize
\caption{\small Scheduler-specific search at 130M/C4/AdamW. Each method uses the common LR grid unless its update rule determines the LR. No non-scheduler training hyperparameter is retuned.}
\label{tab:baseline_tuning}
\begin{tabularx}{\linewidth}{l l X}
\toprule
Method & Selected base LR / WD & Method-specific search $\rightarrow$ selected \\
\midrule
Cosine & $10^{-3}/0.01$ & decay to $0.1\times$ peak \\
WSD & $10^{-3}/0.01$ & stable fraction $\{70,80,90\}\%\rightarrow80\%$; decay starts at $0.9T$ \\
CLR & max $10^{-3}$, base $10^{-4}$ / 0.01 & period $\{1,2,4\}$K $\rightarrow2$K; triangular2 \\
Blockwise LR & $10^{-3}/0.01$ & segments $\{2,3,4\}\rightarrow3$; decay $\{0.5,0.3\}\rightarrow0.5$ \\
Schedule-Free & $10^{-3}/0.01$ & $\beta\in\{0.9,0.95,0.98\}\rightarrow0.9$ \\
Prodigy & adaptive ($d_0=10^{-6}$) / 0.01 & $d_{\rm coef}\in\{0.1,0.5,1.0\}\rightarrow1.0$ \\
MECHANIC & $10^{-3}/0.01$ & number of scales $\{4,6\}\rightarrow6$ \\
AutoLRS & init. $10^{-3}/0.01$ & $K=10$; $\tau_{\mathrm{init}}=1000$; $\tau_{0,\mathrm{init}}=100$; $\tau_{\max}=8000$ \\
AvgLR Replay & inherited / 0.01 & no search; replays SOLAR's mean-LR trajectory \\
SOLAR & inherits Cosine & 60M-only 24-configuration search: $\alpha\in\{0.8,1.0,1.3,1.8\}\rightarrow1.3$; two action-warmup ratios $\rightarrow0.1T$; initial $\log\sigma\in\{-0.7,0,0.7\}\rightarrow0$. Policy LR $3\times10^{-4}$ and PPO clip 0.2 are fixed. \\
\bottomrule
\end{tabularx}
\end{table}

Every configuration in the two-stage screen runs to $0.4T$, after which the top $k=2$ configurations continue to $T$. In a completed 28-point calibration study at 60M, the interim and final rankings have Spearman correlations of $\rho=0.96$ for AdamW and $0.93$ for Muon. Advancing two configurations retains the final winner in both cases; advancing only the top interim configuration would increase the selected final PPL by 0.14.

The SOLAR search runs once at 60M and is reused under both optimizers and all target scales. The later 130M sweep evaluates sensitivity around the selected $\alpha$. The one-time 24-configuration search costs 7.46 wall-clock hours (59.7 GPU-hours) on $8\times$RTX 4090. Five 60M meta-training trajectories cost 1.62 wall-hours (13.0 GPU-hours) for AdamW and 1.99 wall-hours (15.9 GPU-hours) for Muon. Across both optimizers, the pre-submission base searches total 616.91 wall-hours and 4,935.3 GPU-hours. AutoLRS repeats candidate forward/backward segments and costs 1.98--2.00$\times$ its matched baseline; at 1B it uses 113.47 wall-hours versus 56.75 for Cosine.

\subsection{Data, Seeds, Groups, and Recovery State}
\label{app:reproducibility}

The C4 validation stream contains 10M held-out tokens and uses \texttt{T5TokenizerFast} (\texttt{t5-base}, 32K vocabulary). The Pile validation stream contains 1M held-out tokens and uses \texttt{LlamaTokenizerFast}. Loss masks padding tokens. Training seeds are $\{42,52,62\}$ where repeats are available, and the data-shuffle seed is 42. Each trainable tensor forms one controlled group in the dense implementation, giving $G=9L+3=75/111/219/219$ groups from 60M to 1B.

SOLAR maintains two rolling checkpoints at a 1K-update cadence. A rollback restores model weights, optimizer state, LR-scheduler state, mixed-precision state, per-rank random-number-generator state, and data-loader position. The PPO state is not rolled back, so the forced update triggered by the failure is retained. Stale rollouts are discarded, followed by a 1K-update cooldown. A forced $8\times$ LR spike reproduces the reference trajectory after restoration to within $3\times10^{-7}$ and yields identical sample hashes. Checkpoint maintenance adds 0.59\% wall-clock time at 1B and is included in the full-run accounting in Appendix~\ref{app:runtime_overhead}.

\subsection{MoE Model Setup}
\label{app:moe_setup}

Qwen2-MoE~\citep{qwen2technicalreport} is a strong open-source Mixture-of-Experts (MoE) decoder-only Transformer. Compared with the dense Llama~2 models used in our main experiments, its sparse routing mechanism introduces a more challenging optimization landscape, making it a useful testbed for evaluating online learning-rate scheduling under highly non-stationary training dynamics. For our MoE experiments, we train on \textbf{The Pile}~\citep{pile} and build on the Qwen2-MoE codebase released by LITE~\citep{lite}. To align the MoE setting with the Llama~2 experiments, we disable sliding-window attention and use a maximum sequence length of $256$ with batch size $512$. The model is configured to activate $4$ experts per token via a top-$k$ routing strategy ($k=4$), with $32$ experts in total. To encourage stable expert utilization and mitigate representation collapse, we apply a load-balancing auxiliary loss with coefficient $0.01$ and a router $z$-loss with coefficient $0.001$. We train for 100K steps (approximately 13.1B tokens). Table~\ref{tab:qwen2moe_arch} summarizes the full architectural configuration.

\begin{table}[t]
\centering
\caption{\small Architecture of the Qwen2-MoE 1B model.}
\label{tab:qwen2moe_arch}
\renewcommand{\arraystretch}{1.2}
\setlength{\tabcolsep}{8pt}
\begin{tabular}{lcccccccc}
\toprule
Parameter & $d_{\text{model}}$ & $n_{\text{layers}}$ & $n_{\text{heads}}$ & $n_{\text{kv\_heads}}$ & $d_{\text{ff}}$ & $d_{\text{moe\_ff}}$ & $d_{\text{shared\_ff}}$ & $n_{\text{experts}}$ \\
\midrule
Value & 768 & 15 & 12 & 12 & 3072 & 768 & 3072 & 32 \\

\bottomrule
\end{tabular}
\end{table}

\subsection{Optimizer Configurations}
\label{app:opt}

Methods that replace or modify optimizer behavior are described here, whereas external LR schedules are described in Appendix~\ref{app:scheduler_baselines}.

\paragraph{AdamW.}
All Adam-family experiments use the default PyTorch AdamW~\citep{loshchilov2017decoupled} implementation with \(\beta=(0.9, 0.999)\) and \(\epsilon=10^{-8}\). The peak learning rate and weight decay are tuned via grid search as described in Appendix~\ref{app:dense_setup}.

\paragraph{Muon.}
Our Muon implementation follows the standard configuration described by \citet{moonlight_muon}. It uses momentum \(0.95\), Nesterov acceleration, and \(5\) Newton-Schulz orthogonalization steps. The learning rate for matrix-shaped parameters is scaled by \(0.2\sqrt{\max(A,B)}\), where \((A,B)\) are the parameter dimensions. Scalar and embedding parameters are handled by an internal AdamW with \(\beta=(0.9,0.95)\) and \(\epsilon=10^{-8}\).

\paragraph{Prodigy.}
We use the official Prodigy~\citep{mishchenko2024prodigyexpeditiouslyadaptiveparameterfree} implementation without an external learning-rate scheduler. We select $d_{\rm coef}$ using the grid in Table~\ref{tab:baseline_tuning} with weight decay fixed at 0.01; the remaining hyperparameters use their defaults. Because the released implementation is AdamW-based, we report Prodigy only under the AdamW group and omit it from Muon.

\paragraph{Schedule-Free AdamW.}
Schedule-Free AdamW~\citep{defazio2024roadscheduled} is evaluated using the official implementation. We tune the base learning rate on the common validation grid and search $\beta\in\{0.9,0.95,0.98\}$, selecting $\beta=0.9$; weight decay is fixed at 0.01 and the remaining settings follow the official recipe. Because the released implementation is AdamW-based, we report Schedule-Free only under the AdamW group and omit it from Muon.

\subsection{Scheduler Baselines}
\label{app:scheduler_baselines}

We compare SOLAR against the scheduler baselines used in the main paper. All baselines are implemented in the same training codebase and evaluated under the same data, model, batch size, sequence length, and optimizer. Unless otherwise specified, all warmup-decay schedules use the same minimum LR ratio $0.1$. For methods with additional hyperparameters, we tune them using the same validation setup.

\paragraph{Cosine schedule.}
Warmup-Cosine-Decay~(Cosine)~\citep{loshchilov2017sgdrstochasticgradientdescent} uses a linear warmup during the first \(10\%\) of training and then applies cosine annealing to decay the learning rate from \(\eta_{\max}\) to \(0.1\eta_{\max}\) over the remaining steps. We tune the peak learning rate and weight decay within the same validation setting as the other baselines.

\begin{algorithm}[t]
\caption{\small Warmup-Cosine-Decay baseline}
\label{alg:cosine_baseline_appendix}
\KwIn{Base optimizer, peak LR \(\eta_{\max}\), total steps \(T\), warmup ratio \(\rho_w=0.1\)}
\(T_w \leftarrow \lfloor \rho_w T \rfloor\), \(T_c \leftarrow T - T_w\)\;
\For{update step \(t = 0,1,\ldots,T-1\)}{
    \If{\(t < T_w\)}{
        \(\eta_t \leftarrow \eta_{\max}\cdot \frac{t+1}{T_w}\)\;
    }
    \Else{
        \(u \leftarrow t - T_w\)\;
        \(\eta_t \leftarrow 0.1\eta_{\max} + \frac{1}{2}\left(\eta_{\max}-0.1\eta_{\max}\right)\left[1+\cos\left(\pi \frac{u}{T_c}\right)\right]\)\;
    }
    Set the optimizer learning rate to \(\eta_t\) and run one update\;
}
\end{algorithm}

\paragraph{AvgLR Replay baseline.}
This diagnostic baseline is placed immediately after Cosine because it replays SOLAR's scalar average trajectory on the same base scheduling interface, serving as a mechanism test rather than a conventional strong baseline. To construct it, we first record the realized average learning-rate trajectory of a SOLAR run. At each step \(t\), we compute the unweighted mean across controlled parameter groups:
\[
\bar{\eta}_t = \frac{1}{G}\sum_{g=1}^G \eta^{\mathrm{SOLAR}}_{t,g}.
\]
We then replay this scalar trajectory step by step during baseline evaluation and assign the same learning rate to every controlled parameter group:
\[
\eta_{t,g} = \bar{\eta}_t,\quad \forall g.
\]
This baseline uses the SOLAR trajectory directly and does not apply any group-wise weighting.

\paragraph{WSD schedule.}
Warmup-Stable-Decay (WSD)~\citep{hu2024minicpmunveilingpotentialsmall} uses linear warmup during the first \(10\%\) of training, a stable plateau for the next \(80\%\), and a final linear decay during the last \(10\%\). During the decay phase, the learning rate decreases from \(\eta_{\max}\) to \(0.1\eta_{\max}\).

\begin{algorithm}[t]
\caption{\small Warmup-Stable-Decay baseline}
\label{alg:wsd_baseline_appendix}
\KwIn{Base optimizer, peak LR \(\eta_{\max}\), total steps \(T\)}
\(T_w \leftarrow \lfloor 0.1T \rfloor\), \(T_s \leftarrow \lfloor 0.8T \rfloor\), \(T_d \leftarrow T - T_w - T_s\)\;
\For{update step \(t = 0,1,\ldots,T-1\)}{
    \If{\(t < T_w\)}{
        \(\eta_t \leftarrow \eta_{\max}\cdot \frac{t+1}{T_w}\)\;
    }
    \ElseIf{\(t < T_w + T_s\)}{
        \(\eta_t \leftarrow \eta_{\max}\)\;
    }
    \Else{
        \(u \leftarrow t - (T_w + T_s)\)\;
        \(\eta_t \leftarrow \eta_{\max} - \left(\eta_{\max}-0.1\eta_{\max}\right)\frac{u+1}{T_d}\)\;
    }
    Set the optimizer learning rate to \(\eta_t\) and run one update\;
}
\end{algorithm}

\paragraph{CLR schedule.}
CLR~\citep{clr} cyclically varies the learning rate between a lower and an upper bound. We implement CLR as a schedule that can be plugged into the base optimizer used in each experiment, so the training loop uses the same \texttt{optimizer.step()} interface as the other baselines. We use a linear warmup from \(\eta_{\min}\) to \(\eta_{\max}\), followed by a cyclic schedule. Unless otherwise stated, we use \(\eta_{\min}=0.1\eta_{\max}\), the \texttt{triangular2} policy, and a half-cycle length of \(1000\) optimizer update steps. For hyperparameter tuning, we search the peak learning rate, weight decay, and scheduler-specific settings under the same validation setting as the other baselines.

\begin{algorithm}[t]
\caption{\small CLR}
\label{alg:clr_baseline_appendix}
\KwIn{Base optimizer, peak LR \(\eta_{\max}\), lower ratio \(r\), warmup steps \(T_w\), half-cycle length \(s\)}
\(\eta_{\min} \leftarrow r\eta_{\max}\)\;
\For{update step \(t=0,\ldots,T-1\)}{
    \If{\(t<T_w\)}{
        \(\eta_t \leftarrow \eta_{\min} + (\eta_{\max}-\eta_{\min})\frac{t+1}{T_w}\)\;
    }
    \Else{
        \(u \leftarrow t - T_w,\ c \leftarrow \left\lfloor \frac{u}{2s} \right\rfloor\)\;
        \(x \leftarrow \left| \frac{u}{s} - 2c - 1 \right|\), \(a \leftarrow \max(0, 1-x)\)\;
        \(a \leftarrow a / 2^c\)\;
        \(\eta_t \leftarrow \eta_{\min} + (\eta_{\max}-\eta_{\min})a\)\;
    }
    Set the optimizer learning rate to \(\eta_t\) and run one update\;
}
\end{algorithm}

\paragraph{Blockwise LR.}
Blockwise LR~\citep{blockwise} assigns block-type-specific learning rates based on the transformer sharpness disparity principle. We use the official AdamW-based implementation, tune the peak learning rate on the common validation grid, and search the number of segments over $\{2,3,4\}$ and the decay factor over $\{0.5,0.3\}$, selecting three segments and a decay factor of 0.5; weight decay is fixed at 0.01. Because the released implementation is AdamW-based, we report Blockwise LR only under the AdamW group and omit it from Muon.

\paragraph{AutoLRS schedule.}
AutoLRS~\citep{autoLRS} selects a piecewise-constant learning rate online through repeated short-horizon candidate evaluation. At the beginning of each stage, AutoLRS buffers the next \(\tau\) update batches, checkpoints the model, optimizer, and random-number-generator states, and evaluates \(K\) candidate learning rates. Each candidate is trained for only \(\tau_0\) update steps from the same checkpoint and on the same buffered data. The short-horizon loss curve is then summarized by an exponential forecaster and used as the objective for Bayesian optimization. After the \(K\) trials, AutoLRS restores the original checkpoint and trains the real model for the full \(\tau\)-step stage using the candidate LR with the lowest predicted loss.

We use \(K=10\), \(\kappa=1000\) (the LCB exploration coefficient), \(\tau_{\mathrm{init}}=1000\), \(\tau_{0,\mathrm{init}}=100\), and \(\tau_{\max}=8000\). The stage length doubles after each stage until it reaches \(\tau_{\max}\), and \(\tau_0\) is set to one tenth of the current stage length. During early stages, the objective is the training loss curve; once the stage length reaches \(\tau_{\max}\), we switch to a fixed validation subset of 4 pre-cached C4 validation batches. AutoLRS assigns the selected LR to all parameter groups.

\begin{algorithm}[t]
\caption{\small AutoLRS}
\label{alg:autolrs_baseline_appendix}
\KwIn{Base optimizer, LR interval \([\eta_{\min},\eta_{\max}]\), number of BO trials \(K\), stage length \(\tau\), probe length \(\tau_0\)}
\While{training is not finished}{
    Buffer the next \(\tau\) update batches and save model, optimizer, and RNG states\;
    Reset the Bayesian optimizer for the current stage\;
    \For{\(i=1,\ldots,K\)}{
        Suggest candidate \(\eta_i\) by LCB acquisition with exploration coefficient \(\kappa\) over \([\log \eta_{\min}, \log \eta_{\max}]\)\;
        Restore the saved state\;
        Train for \(\tau_0\) buffered update steps with global LR \(\eta_i\)\;
        Record the short-horizon loss curve\;
        Fit the exponential forecaster and predict the stage loss \(\hat{\ell}_i\)\;
        Update the GP with \((\log \eta_i, \hat{\ell}_i)\)\;
    }
    Select \(i^\star = \arg\min_i \hat{\ell}_i\)\;
    Restore the saved state and train the real model for \(\tau\) steps with global LR \(\eta_{i^\star}\)\;
    Increase \(\tau\) up to \(\tau_{\max}\) and set \(\tau_0 \leftarrow \max(1,\tau/10)\)\;
}
\end{algorithm}

\paragraph{MECHANIC schedule.}

MECHANIC~\citep{mechanic} learns a single model-wide scale that rescales the cumulative update direction produced by the base optimizer. The method maintains a reference copy of the initial parameters, a cumulative base-update direction, and a collection of tuner states that determine the next global scale. At each step, MECHANIC first applies one update of the base optimizer, then computes a scalar feedback signal from the current gradient, the old cumulative update direction, and global norms. This feedback is used to update the tuner states and obtain a new scale \(S_{t+1}\). Finally, the parameters are reparameterized around the reference point as
\[
x_{t+1} = x_{\mathrm{ref}} + S_{t+1}\Delta_{t+1}.
\]

We use the official MECHANIC implementation under the same training setup as the corresponding baseline. The base learning rate and weight decay follow the common validation search, and Table~\ref{tab:baseline_tuning} reports the number-of-tuners search. The remaining MECHANIC settings follow the official implementation.

\subsection{Hardware}
All dense and Qwen2-MoE experiments up to 1B parameters run on $8\times$NVIDIA RTX 4090 GPUs. The DeepSeek-V2-style 3B MoE experiment runs on $32\times$NVIDIA A800 80GB GPUs. All dense and 1B MoE wall-clock and overhead measurements use the RTX 4090 system; the 3B MoE timing uses the A800 system.

\section{Additional Ablation and Experimental Results}
\label{app:additional_ablations}

Unless otherwise specified, the following ablations use the seed-52 run of a 130M-parameter Llama~2 model trained on C4 with AdamW under Appendix~\ref{app:dense_setup}. They isolate the state, action, reward, control granularity, safety, and transfer choices used by SOLAR.

\subsection{Positioning among Learned LR Controllers}
\label{app:learned_lr_positioning}

Table~\ref{tab:learned_lr_positioning} distinguishes the object learned by each controller and the runs used to acquire it. Prior work establishes RL-based global scaling, state-conditioned actions, transfer, and layer-wise control. SOLAR targets a different acquisition regime: the controller learns during the same full autoregressive LLM pretraining trajectory that it must improve.

\begin{table}[H]
\centering
\footnotesize
\caption{\small Positioning among representative learned LR controllers. ``Same live run'' means that policy updates occur inside the final run being improved.}
\label{tab:learned_lr_positioning}
\begin{tabularx}{\linewidth}{@{}>{\raggedright\arraybackslash}p{0.22\linewidth}>{\raggedright\arraybackslash}p{0.18\linewidth}>{\raggedright\arraybackslash}p{0.25\linewidth}>{\raggedright\arraybackslash}X@{}}
\toprule
Method & Learned object & Acquisition & Evaluated setting \\
\midrule
Daniel et al. (2016) & global step size & repeated episodes & small networks \\
Xu et al. (2017) & global absolute LR & actor--critic with resets & vision \\
Xu et al. (2019) & global LR profile & past complete histories & Fashion-MNIST, CIFAR-10 \\
GNS (2022) & global graph policy & separate target episodes & vision, GLUE \\
Subramanian et al. (2023) & PPO LR schedule & separate controller training & MNIST, CIFAR-100 \\
GANNO (2023) & layer-wise absolute LR & separate environments & vision \\
SOLAR-online & bounded group residual & same live run & autoregressive LLM pretraining \\
SOLAR-frozen & bounded group residual & source proxy runs & dense scale/corpus transfer \\
\bottomrule
\end{tabularx}
\end{table}

\subsection{Repeated Final-Checkpoint Results}
\label{app:seed_results}

Table~\ref{tab:seed_results} lists final PPL for every repeated run and reports the mean across available seeds. Hyperparameters remain fixed after selection. The two SOLAR-frozen rows reuse one source-trained policy across all three target seeds.

\begin{table}[H]
\centering
\footnotesize
\caption{\small Final-checkpoint PPL by seed. Seed 42 is the selection seed for online and baseline configurations; seeds 52 and 62 use the selected configuration. SOLAR-frozen reuses one source-trained policy across all target seeds.}
\label{tab:seed_results}
\begin{tabular}{llcccc}
\toprule
Setting & Method & Seed 42 & Seed 52 & Seed 62 & Mean PPL \\
\midrule
60M AdamW & Cosine & 30.71 & 30.49 & 30.34 & 30.51 \\
 & WSD & 29.63 & 29.80 & 29.91 & 29.78 \\
 & SOLAR-online & 28.99 & 28.92 & 29.14 & $\mathbf{29.02}$ \\
\midrule
130M AdamW & Cosine & 24.41 & 24.52 & 24.57 & 24.50 \\
 & WSD & 24.07 & 23.96 & 23.90 & 23.98 \\
 & SOLAR-online & 22.95 & 22.79 & 22.84 & $\mathbf{22.86}$ \\
 & SOLAR-frozen & 22.96 & 22.41 & 22.90 & $\mathbf{22.76}$ \\
\midrule
130M Muon & Cosine & 22.61 & 22.55 & 22.44 & 22.53 \\
 & SOLAR-online & 22.11 & 21.99 & 22.06 & $\mathbf{22.05}$ \\
 & SOLAR-frozen & 22.24 & 21.87 & 22.16 & $\mathbf{22.09}$ \\
\midrule
350M AdamW & Cosine & 18.42 & 18.31 & 18.47 & 18.40 \\
 & SOLAR-online & 17.19 & 17.23 & 17.12 & $\mathbf{17.18}$ \\
\midrule
1B AdamW & Cosine & 16.44 & 16.52 & -- & 16.48 \\
 & SOLAR-online & 14.90 & 14.83 & -- & $\mathbf{14.87}$ \\
\bottomrule
\end{tabular}
\end{table}

SOLAR is lower in every paired 350M and 1B run. The paired gaps are 1.23/1.08/1.35 PPL at 350M and 1.54/1.69 at 1B. At 130M, online/frozen mean PPLs are 22.86/22.76 under AdamW and 22.05/22.09 under Muon.

\subsection{Online Acquisition and Frozen Reuse}
\label{app:online_frozen_modes}

SOLAR supports in-run acquisition and amortized reuse of the same controller design. SOLAR-online updates the policy inside the target run. SOLAR-frozen reuses a policy acquired over $K$ full-length 60M online runs and performs no target PPO updates. Freezing fixes all source-policy parameters, including the final learned shared $\log\sigma$, not the action sequence: the controller continues to recompute group-wise actions from each target state.

\begin{table}[H]
\centering
\caption{\small Source-side acquisition and 130M/C4/AdamW frozen reuse. Every target evaluation uses seed 52. The $K=0$ row keeps the policy at its random initialization.}
\label{tab:source_trajectory_sweep}
\begin{tabular}{cc}
\toprule
Full-length 60M source runs $K$ & Frozen final PPL \\
\midrule
0 & 25.04 \\
1 & 23.26 \\
2 & 22.89 \\
3 & 22.58 \\
5 & \textbf{22.41} \\
\bottomrule
\end{tabular}
\end{table}

Frozen reuse improves as the policy is acquired over more source runs, moving from 25.04 PPL at $K=0$ to 22.41 at $K=5$. Each source run is a complete online pretraining trajectory.

\begin{table}[H]
\centering
\caption{\small Online acquisition and frozen reuse across source--target settings. Rows with C4 targets use seed 52; the Pile-target row reports the mean of seeds 42 and 52.}
\label{tab:online_frozen_conditions}
\begin{tabularx}{\linewidth}{@{}>{\raggedright\arraybackslash}Xccc@{}}
\toprule
Source $\rightarrow$ target & Cosine & SOLAR-frozen & SOLAR-online \\
\midrule
60M/C4/AdamW $\rightarrow$ 130M/C4/AdamW & 24.52 & \textbf{22.41} & \textbf{22.79} \\
60M/C4/Muon $\rightarrow$ 130M/C4/Muon & 22.55 & \textbf{21.87} & \textbf{21.99} \\
60M/C4/AdamW $\rightarrow$ 130M/Pile/AdamW & 13.96 & \textbf{13.21} & \textbf{12.99} \\
60M/C4/AdamW $\rightarrow$ 130M/C4/Muon & 22.55 & \textbf{22.49} & \textbf{21.99} \\
\bottomrule
\end{tabularx}
\end{table}

Both SOLAR modes remain below Cosine in the optimizer-matched rows. Frozen reuse reaches the lower PPL on the matched C4 targets, while online acquisition reaches the lower PPL under the corpus and optimizer changes.

\subsection{Matched Learned-Controller Comparisons}
\label{app:novelty_controls}

The 130M/C4/AdamW controls in Table~\ref{tab:novelty_controls} share 20K updates, weight decay 0.01, and a tuned peak LR of $10^{-3}$ where a base schedule is used. N1--N3 are design-axis probes; N4 is an independent LLM adaptation of GANNO with its own controller grid. Each control is tuned on its own search space.

\paragraph{Implementations and tuning protocol.}
N1 and N2 use the same global, state-conditioned PPO controller with $G=1$. N1 applies a bounded multiplier that re-anchors to the shared base schedule at every update; N2 compounds each multiplier on the preceding LR. N3 retains the anchored group-wise interface but replaces PPO with a myopic Baydin-style hypergradient rule. These three controls are trained directly in the 130M target run, and each is searched on its own controller grid. N4 follows the GANNO control structure in a separate implementation: a shared recurrent IPPO actor--critic serves 15 module-level agents, chooses among nine categorical actions on cumulative absolute LRs around $\eta^*=10^{-3}$ every 50 updates, and uses counterfactual difference rewards. Its policy is meta-trained at 60M, frozen, and evaluated at 130M with no target PPO updates; its controller is searched on its own grid. N1--N3 isolate individual design axes, whereas N4 is a complete competing controller adapted to the LLM setting.

\begin{table}[H]
\centering
\footnotesize
\caption{\small Matched controls that isolate the SOLAR control structure. The matched SOLAR/N1/N2 comparisons use seeds 42 and 52; each arrow gives the mean. Three-run static-baseline summaries include the sample standard deviation.}
\label{tab:novelty_controls}
\begin{tabularx}{\linewidth}{@{}>{\raggedright\arraybackslash}p{0.35\linewidth}>{\raggedright\arraybackslash}p{0.32\linewidth}>{\raggedright\arraybackslash}X@{}}
\toprule
Method and control structure & Per-seed PPL $\rightarrow$ summary & Isolated axis \\
\midrule
SOLAR-online: anchored group residual, $G=111$ & $22.95/22.79\rightarrow\mathbf{22.87}$ & full method \\
N1 Global-PPO-Residual: anchored, $G=1$ & $23.68/23.80\rightarrow23.74$ & group granularity \\
WSD: tuned static base & \makecell[l]{$24.07/23.96/23.90$\\$\rightarrow23.98\pm0.09$} & learned control \\
N3 Group-MHD: anchored, group-wise, non-RL & $24.31/24.21\rightarrow24.26$ & learned feedback \\
Cosine: tuned static base & \makecell[l]{$24.41/24.52/24.57$\\$\rightarrow24.50\pm0.08$} & learned control \\
N4 GANNO-IPPO: layer-wise absolute LR & $24.89/25.01\rightarrow24.95$ & residual design \\
N2 Global-PPO-Recursive: unanchored, $G=1$ & $27.30/26.88\rightarrow27.09$ & anchoring and bounds \\
\bottomrule
\end{tabularx}
\end{table}

N1 and N2 use the same global, state-conditioned PPO interface. On the common seeds 42 and 52, re-anchoring each multiplier to the base and bounding the action improves 27.09 to 23.74, a 3.35-PPL change. Moving from one global action to $G=111$ group actions further improves PPL to 22.87; the third SOLAR run gives a three-run mean of $22.86\pm0.08$ in Table~\ref{tab:seed_results}. N3 shows that a group-wise residual without a learned state-conditioned loop does not recover the same gain. N4 provides a layer-wise RL controller adapted to the same LLM setting. Under 60M-to-130M frozen-policy transfer, N4 reaches 24.95 PPL and SOLAR-frozen reaches a three-run mean of 22.76, a 2.19-PPL gap between the two LLM controller adaptations. Within SOLAR, replacing the residual action with a target-trained absolute-LR action changes 22.79 to 24.74, a 1.95-PPL gap. With the SOLAR architecture held fixed, untrained, near-deterministic, and state-blind policies reach 25.04, 25.03, and 25.21; AvgLR Replay reaches 25.61 at the final checkpoint. Together, these controls isolate the contribution of the anchored residual interface, state-conditioned learning, and group-wise actions.

\paragraph{Failure-mode diagnostics.}
The N1/N2 pairing exposes a slow failure that a loss-spike safeguard does not detect. In an $\alpha=2.6$ stress run ($n=1$), N2 produces no loss spike: its recursively updated LR drifts to a $5.7\times$ mean multiplier and finishes at 79.58 PPL. N3 exposes a second failure mode. Moving its single hypergradient coefficient by one decade changes PPL at the $0.4T$ screen from 31.05 to 47.31, showing sharp sensitivity to its local update scale.

\paragraph{Action-bound sensitivity.}
The main setting $\alpha=1.3$ is inherited from the 60M search; Table~\ref{tab:alpha_sensitivity} evaluates sensitivity around this setting at 130M. Performance remains better than Cosine over a twofold range, and the Circuit-Breaker first activates at $\alpha=1.8$.

\begin{table}[H]
\centering
\caption{\small Sensitivity to the residual action bound on 130M/C4/AdamW.}
\label{tab:alpha_sensitivity}
\begin{tabular}{lccccc}
\toprule
$\alpha$ & 0.8 & 1.0 & 1.3 & 1.6 & 1.8 \\
\midrule
Final PPL & 23.42 & 23.05 & \textbf{22.79} & 22.94 & 23.61 \\
CB activation & no & no & no & no & yes \\
\bottomrule
\end{tabular}
\end{table}

\subsection{Frozen-Policy Robustness and \texorpdfstring{$\mu$P}{muP} Compatibility}
\label{app:transfer_robustness}

\paragraph{Target base-LR sweep.}
We freeze a 60M/C4 policy and apply it with no target PPO updates. Table~\ref{tab:mistuned_transfer} varies the target peak LR over a fourfold window. Each cell is a single seed-52 run; the experiment measures sensitivity to the supplied base LR rather than run-to-run variance.

\begin{table}[H]
\centering
\caption{\small Frozen residual-policy transfer under misspecified target base LRs.}
\label{tab:mistuned_transfer}
\begin{tabular}{ccccc}
\toprule
Base peak & 130M Cosine & 130M frozen & 350M Cosine & 350M frozen \\
\midrule
$0.5\times$ & 25.05 & \textbf{22.72} & 18.79 & \textbf{17.61} \\
$1.0\times$ & 24.52 & \textbf{22.41} & 18.31 & \textbf{17.37} \\
$2.0\times$ & diverged @4.1K & \textbf{22.68} & diverged @6.8K & \textbf{17.96} \\
\bottomrule
\end{tabular}
\end{table}

The frozen policy beats the same-base Cosine run in all six settings and completes both $2\times$ runs. At 130M, the negative-action fraction changes from 0.29 to 0.47 to 0.73 as the base rises from $0.5\times$ to $2\times$, showing bidirectional state-conditioned correction rather than a fixed positive multiplier.

The main-table protocol searches LR/WD at each target scale, at costs of 14.02, 35.66, and 227.65 wall-clock hours for 130M, 350M, and 1B. Freezing reuses the resulting base schedule and removes target PPO training and target-scale search over SOLAR-specific hyperparameters. The avoided controller-side work is 0.9$\times$, 2.7$\times$, and 2.7$\times$ the corresponding LR/WD-search cost.

\paragraph{$\mu$P width transfer.}
We combine SOLAR with maximal-update parameterization ($\mu$P)~\citep{tensorprogramsvtuning}. A 71M base and 130M target differ only in hidden width; both use 12 layers, head dimension 64, $G=111$, sequence length 256, global batch 512, 20K updates, weight decay 0.01, and 10\% warmup. A three-point search at 71M selects $\eta^*=2\times10^{-3}$. $\mu$P transfers per-tensor target LRs from fan-in: $2.000\times10^{-3}$ for embeddings/RMSNorm and approximately $1.333\times10^{-3}$ for attention, MLP, and readout tensors. No LR/WD search occurs at 130M.

\begin{table}[H]
\centering
\caption{\small SOLAR on a directly searched standard-parameterization (SP) base and a $\mu$P-transferred base at 130M/C4/AdamW. For the two-run $\mu$P settings, values before the arrow are per-seed final PPL and the arrow gives their mean.}
\label{tab:mup_solar}
\begin{tabular}{lcc}
\toprule
Method & Searched SP base & $\mu$P-transferred base \\
\midrule
Cosine & $24.50\pm0.08$ ($n=3$) & $24.44/24.53\rightarrow24.49$ \\
SOLAR-online & $22.86\pm0.08$ ($n=3$) & $22.87/22.94\rightarrow\mathbf{22.91}$ \\
SOLAR-frozen & 22.41 ($n=1$) & $22.62/22.74\rightarrow\mathbf{22.68}$ \\
\bottomrule
\end{tabular}
\end{table}

$\mu$P+Cosine reaches 24.49 PPL, compared with 24.50 for the searched SP base. SOLAR improves each paired $\mu$P run by 1.57--1.59 PPL online and 1.79--1.82 frozen. The frozen configuration performs neither LR/WD search nor PPO updates at the 130M target. It inherits $\alpha=1.3$, action warmup $0.1T$ ($=2$K updates), the source policy's final learned $\log\sigma$, and all reward coefficients. The online policy's negative-action fraction is 0.44 under $\mu$P and 0.47 under SP.

We validate the $\mu$P implementation by checking activation-scale consistency across widths. Across hidden widths 256--1024, the mean absolute activation at each layer changes by less than 6\% under $\mu$P; under standard parameterization, the width-1024 value is approximately $2.1\times$ the width-256 value.

\paragraph{Corpus shift.}
We also train 130M models on The Pile with the same architecture and token budget. The Cosine peak LR is re-searched on The Pile over $\{5\times10^{-4},10^{-3},3\times10^{-3}\}$; $10^{-3}$ is selected and $3\times10^{-3}$ diverges at 3.4K updates. Table~\ref{tab:pile_transfer} reports the two final PPL values for each method.

\begin{table}[H]
\centering
\caption{\small Cross-corpus evaluation on 130M/The Pile/AdamW.}
\label{tab:pile_transfer}
\begin{tabular}{lcccc}
\toprule
Method & Seed 52 & Seed 42 & Mean & Gain vs. Cosine \\
\midrule
Cosine & 13.90 & 14.02 & 13.96 & -- \\
SOLAR-online & 12.94 & 13.03 & \textbf{12.99} & 0.97 (6.9\%) \\
SOLAR-frozen (60M/C4 policy) & 13.18 & 13.24 & \textbf{13.21} & 0.75 (5.4\%) \\
\bottomrule
\end{tabular}
\end{table}

The online policy preserves its C4 relative gain under the corpus shift. On The Pile, online has a 0.22 lower mean PPL than frozen across seeds 42 and 52; on the seed-52 C4 run, frozen is 0.38 PPL lower than online. Both modes remain below their matched Cosine baselines.

\subsection{Static and Open-Loop Replay Controls}
\label{app:open_loop_replays}

We derive replay controls from a completed 130M/C4/AdamW/seed-52 SOLAR-online run. The static group profile averages each of the $G=111$ multipliers over 20K updates and normalizes their cross-group mean to one. The time-varying replay applies the recorded multiplier for every group and update. The level-corrected scalar replay rescales SOLAR's step-wise mean LR before replay. Each replay runs open-loop without a policy, state input, PPO update, or Circuit-Breaker. A five-point global rescaling search for the static profile selects $c=1.0$ and retains its 24.19 final PPL.

\begin{table}[H]
\centering
\caption{\small Replay controls on 130M/C4/AdamW with seed 52. All entries report final PPL.}
\label{tab:replay_controls}
\begin{tabular}{lcccc}
\toprule
Method & Group-wise & Time-varying & Live state & Final PPL \\
\midrule
Cosine & -- & base only & -- & 24.52 \\
AvgLR Replay & -- & $\checkmark$ & -- & 25.61 \\
Level-corrected scalar replay & -- & $\checkmark$ & -- & 24.63 \\
Static normalized group profile & $\checkmark$ & -- & -- & 24.19 \\
SOLAR-online & $\checkmark$ & $\checkmark$ & $\checkmark$ & \textbf{22.79} \\
\bottomrule
\end{tabular}
\end{table}

AvgLR Replay and the level-corrected scalar replay reach 25.61 and 24.63 PPL. Static group allocation improves Cosine from 24.52 to 24.19 PPL, while SOLAR-online reaches 22.79.

\begin{table}[H]
\centering
\footnotesize
\caption{\small SOLAR-derived replays across seeds, base LRs, and corpora. Both replay profiles come from the seed-52 C4 run at the selected base LR. The Pile row uses seed 52; ``--'' denotes a setting not evaluated.}
\label{tab:replay_transfer}
\begin{tabularx}{\linewidth}{@{}>{\raggedright\arraybackslash}Xccccc@{}}
\toprule
Setting & Cosine & Static group & Time-varying group & SOLAR-frozen & SOLAR-online \\
\midrule
C4, seed 42, $1.0\times$ base & 24.41 & 24.29 & 23.94 & 22.96 & \textbf{22.95} \\
C4, seed 52, $0.5\times$ base & 25.05 & 24.83 & 24.47 & \textbf{22.72} & -- \\
C4, seed 52, $2.0\times$ base & div. @4.1K & div. @3.8K & div. @4.4K & \textbf{22.68} & -- \\
Pile, seed 52, $1.0\times$ base & 13.90 & 13.71 & 13.55 & 13.18 & \textbf{12.94} \\
\bottomrule
\end{tabularx}
\end{table}

The learned policies remain below both replay controls across the seed, base-LR, and corpus shifts. At the $2.0\times$ base LR, SOLAR-frozen completes training at 22.68 PPL, while Cosine and both open-loop replays diverge.

\subsection{Importance of Global and Local State Features}
\label{app:ablation_state}

To evaluate our state formulation, we compare the \textbf{Full state} (the complete 10-dimensional state described in Appendix~\ref{app:state_impl}) against two variants:
\begin{itemize}
    \item \textbf{Global-only state}: Omits per-group statistics, relying solely on the 4 shared global features (e.g., training step, global loss).
    \item \textbf{Local-only state}: Omits macro-phase indicators, relying solely on the 6 per-group local features (e.g., layer-wise gradient norms).
\end{itemize}

Combining global and local features gives the lowest final PPL in Table~\ref{tab:state_representation}: 22.79, compared with 23.11 for global-only and 23.57 for local-only.
\begin{table}[H]
\centering
\caption{\small Ablation on state representation for the 130M model.}
\label{tab:state_representation}
\begin{tabular}{lc}
\toprule
State Variant    & Final Eval PPL ($\downarrow$) \\
\midrule
Full state (Ours) & 22.79 \\
Global-only        & 23.11 \\
Local-only       & 23.57 \\
\bottomrule
\end{tabular}
\end{table}

\subsection{Stochastic Exploration vs. Near-Deterministic Scheduling}
\label{app:ablation_action_space}

SOLAR relies on stochastic exploration to gather informative credit assignment signals under noisy, delayed loss feedback. To quantify the contribution of stochastic exploration, we fix the policy standard deviation to a near-zero constant \(\sigma_{\mathrm{det}}\) throughout training while retaining the same PPO update and residual action parameterization:
\[
\begin{aligned}
u_{t,g}&=\mu_{t,g}+\sigma_{\mathrm{det}}\epsilon_{t,g},
&\epsilon_{t,g}&\sim\mathcal N(0,1),
&\sigma_{\mathrm{det}}&\approx0,\\
\widetilde a_{t,g}&=\tanh(u_{t,g}),
&a_{t,g}&=\operatorname{clip}(\widetilde a_{t,g},-1+10^{-4},1-10^{-4}).
\end{aligned}
\]
The resulting actions are effectively determined by the policy mean.
All other components---the residual LR modulation (\eqref{eq:residual_lr}), action-scale warmup schedule, PPO surrogate objective, value function, entropy regularization, and network architecture---remain identical to the stochastic policy.

Suppressing stochasticity increases final PPL from 22.79 to 25.03 in Table~\ref{tab:action_space}. The result is consistent with stochastic sampling improving exploration under delayed feedback.

\begin{table}[H]
\centering
\caption{\small Comparison of stochastic exploration versus near-deterministic scheduling (130M model).}
\label{tab:action_space}
\begin{tabular}{lc}
\toprule
Policy Formulation & Final Eval PPL ($\downarrow$) \\
\midrule
Stochastic (Ours)  & 22.79 \\
Near-deterministic  & 25.03 \\
\bottomrule
\end{tabular}
\end{table}

\subsection{Residual LR Modulation versus Direct Absolute LR Prediction}
\label{subsec:ablation_curriculum}

A natural baseline for RL-based learning-rate scheduling is to predict the absolute learning rate directly at each step. We instantiate this idea as a \textbf{Direct-Abs} baseline. Specifically, the scheduler first samples an unconstrained latent action,
\[
u_{t,g} \sim \mathcal{N}(\mu_{t,g}, \sigma^2),
\qquad
\widetilde a_{t,g}=\tanh(u_{t,g}),\qquad
a_{t,g}=\operatorname{clip}(\widetilde a_{t,g},-1+10^{-4},1-10^{-4})
\]
where \(\mu_{t,g}\) is the scheduler output and the scalar \(\sigma\) is a shared learnable parameter controlling action stochasticity. The squashed action \(a_{t,g}\) is then mapped to an absolute learning rate in log-space:
\[
\log \eta_{t,g}
=
\log \eta_{\min}
+
\frac{a_{t,g}+1}{2}
\left(
\log \eta_{\max} - \log \eta_{\min}
\right),
\]
and the final learning rate is obtained by exponentiation,
\[
\eta_{t,g} = \exp\!\left(\log \eta_{t,g}\right).
\]
This parameterization keeps the action bounded and well-defined, while making the scheduler operate directly on the absolute learning-rate scale. We set \(\eta_{\min} = 0.1 \eta_{\max}\) and select \(\eta_{\max}\) from the same seven-point LR grid used for the other schedulers.

Despite this tuning, Direct-Abs remains brittle in practice. Because the scheduler is responsible for both the absolute scale and the temporal evolution of the learning rate from the very beginning of training, early miscalibrated actions can induce overly aggressive updates, destabilize optimization, and even trigger pronounced loss spikes. Once such disruptions occur in the early stage, they are often difficult to recover from, which ultimately degrades final performance.

SOLAR therefore adopts residual LR modulation on top of a base scheduler. Specifically, the scheduler predicts a bounded multiplicative adjustment:

\[ \eta_{t,g} = \eta_{t,g}^{\mathrm{base}} \cdot \exp(\alpha_t a_{t,g}) \]

where \(a_{t,g}\) is the scheduler action and \(\alpha_t\) controls the residual modulation strength. To further stabilize early training, we linearly warm up the residual scale \(\alpha_t\) from 0 to \(\alpha\), which limits the magnitude of residual perturbations before the scheduler becomes reliable.

The combined residual-and-warmup design reaches 22.79 final PPL, compared with 24.74 for Direct-Abs under the matched LR search (Table~\ref{tab:curriculum_residual}).

\begin{table}[H]
\centering
\caption{\small Ablation on residual LR modulation versus direct absolute LR prediction (130M model).}
\label{tab:curriculum_residual}
\begin{tabular}{lc}
\toprule
Design & Final Eval PPL (\(\downarrow\)) \\
\midrule
SOLAR (Residual + Warmup) & 22.79 \\
Direct-Abs & 24.74 \\
\bottomrule
\end{tabular}
\end{table}

\subsection{Robustness to Suboptimal Base Learning Rates}
\label{app:lr_robustness}

Figure~\ref{fig:lr_sensitivity_a} complements the matched $0.5\times$/$1\times$/$2\times$ transfer study in Appendix~\ref{app:transfer_robustness} with a wider sensitivity sweep on the 130M AdamW setting. We vary the peak base LR from $10^{-4}$ to $10^{-2}$ for Cosine, WSD, and SOLAR with a Cosine base.

Cosine and WSD deteriorate once the peak LR exceeds their selected operating region and diverge at the largest values. SOLAR remains trainable over a wider observed interval and, at high base LRs, uses negative residual actions to reduce the effective LR. Across the evaluated sweep, SOLAR tolerates substantial base-LR misspecification and actively compensates when the supplied LR is too large.

\begin{figure}[htbp]
    \centering
    \begin{subfigure}[t]{0.49\textwidth}
        \centering
        \includegraphics[width=\textwidth]{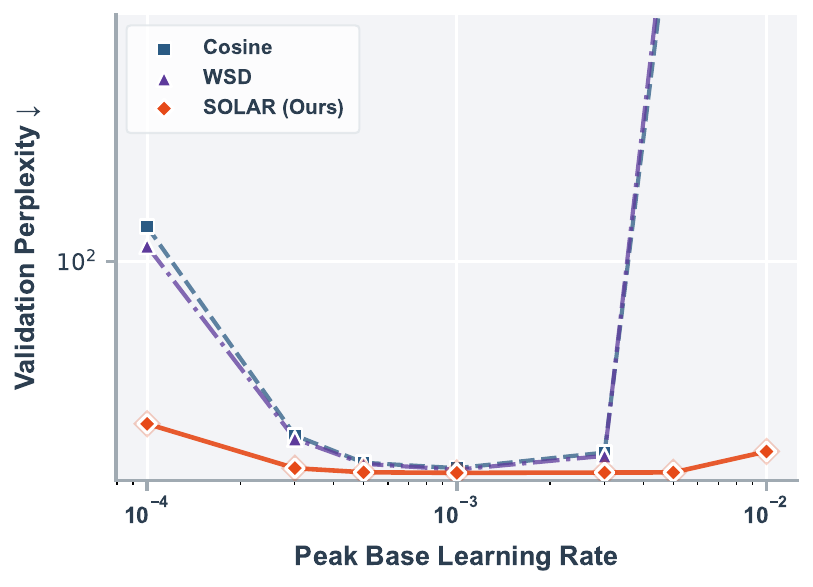}
        \caption{Final PPL across peak base LRs on the 130M AdamW setting. Missing high-LR points denote diverged runs.}
        \label{fig:lr_sensitivity_a}
    \end{subfigure}
    \hfill
    \begin{subfigure}[t]{0.49\textwidth}
        \centering
        \includegraphics[width=\textwidth]{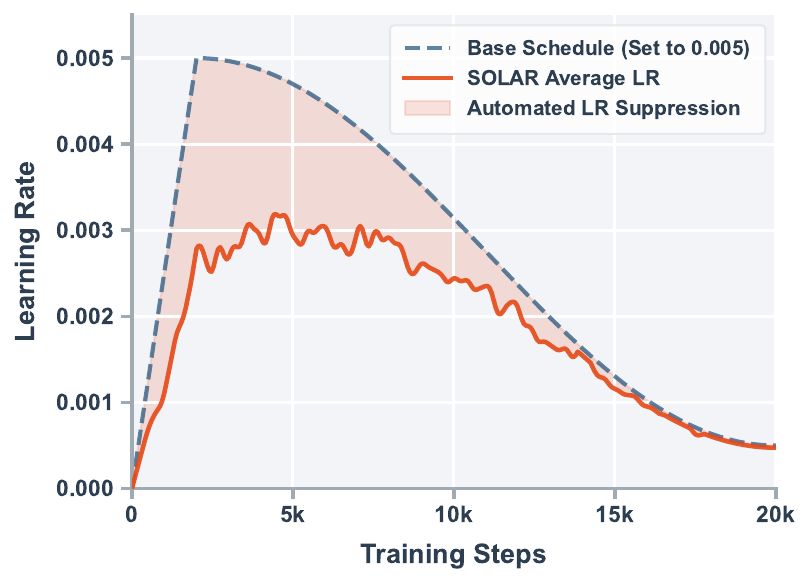}
        \caption{At a peak base LR of \(0.005\), negative residual actions reduce the effective LR below the supplied base schedule.}
        \label{fig:lr_sensitivity_b}
    \end{subfigure}
    \caption{Sensitivity analysis and mechanism illustration for SOLAR.}
    \label{fig:lr_sensitivity}
\end{figure}

\textbf{Observed response.} Figure~\ref{fig:lr_sensitivity_b} records the controller response at a peak base LR of $0.005$. As gradient statistics worsen, the policy outputs $a_{t,g}<0$ for affected groups. The mapping $\eta_{t,g}=\eta_{t,g}^{\mathrm{base}}\exp(\alpha_t a_{t,g})$ then lowers their effective LRs without replacing the long-horizon base profile. This trace provides a direct mechanism for the wider stable interval observed in Figure~\ref{fig:lr_sensitivity_a}.

\subsection{Efficacy of the Automated Circuit-Breaker Mechanism}
\label{app:circuit_breaker_exp}

The Circuit-Breaker handles rare optimization anomalies by restoring a complete training state. None of the main-table or MoE runs triggers it. We therefore validate recovery separately with a stress test.

\textbf{Experimental Setup.} We stress-test the Circuit-Breaker on the 130M model by increasing the action bound to $\alpha=1.8$, which makes large LR multipliers more likely.

\begin{figure}[htbp]
    \centering
    \begin{subfigure}[t]{0.49\textwidth}
        \centering
        \includegraphics[width=\textwidth]{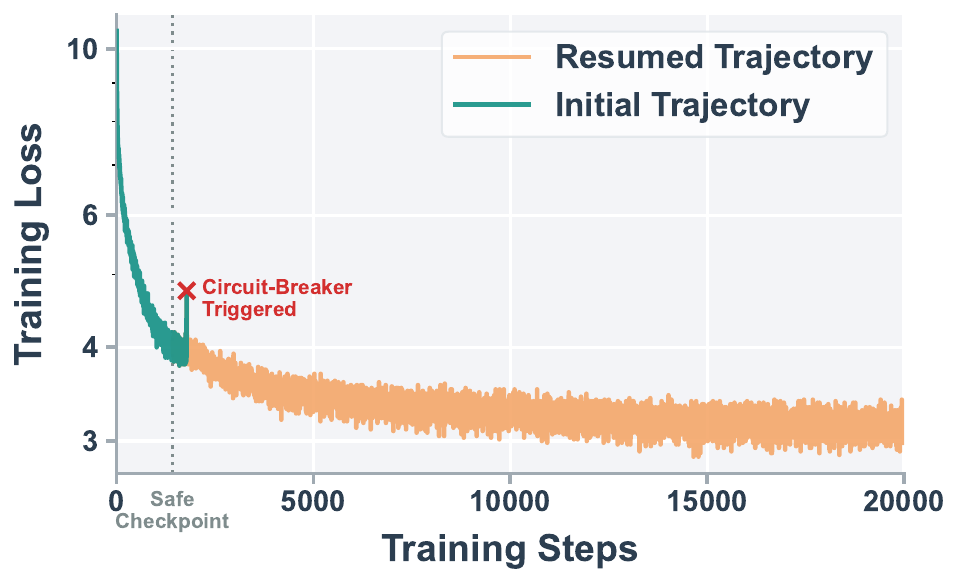}
        \caption{\small Training loss trajectories.}
        \label{fig:circuit_breaker_loss}
    \end{subfigure}
    \hfill
    \begin{subfigure}[t]{0.49\textwidth}
        \centering
        \includegraphics[width=\textwidth]{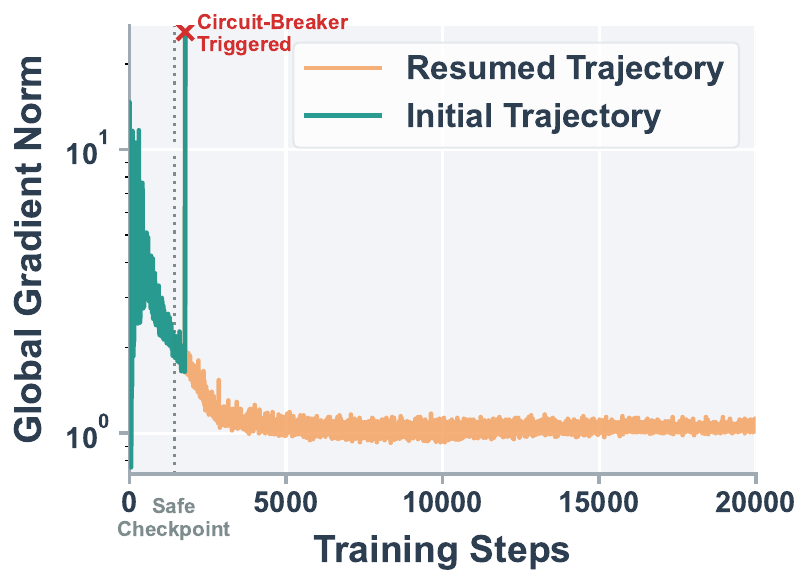}
        \caption{\small Global gradient norm trajectories.}
        \label{fig:circuit_breaker_gradnorm}
    \end{subfigure}
    
    \caption{\small Demonstration of the automated Circuit-Breaker mechanism under an aggressive exploration stress test (\(\alpha = 1.8\)). In both panels, the teal line represents the initial training rollout, which experiences a severe spike and triggers the Circuit-Breaker (marked by the red cross). The orange line represents the automatically resumed trajectory, which restarts from the last safe checkpoint (indicated by the vertical dotted line) and successfully completes the training process.}
    \label{fig:circuit_breaker}
\end{figure}

\textbf{Results and Analysis.} As illustrated in Figure~\ref{fig:circuit_breaker}, the initial training rollout (teal line) proceeds normally until the agent samples an overly aggressive learning rate multiplier, causing a sudden and severe spike in both the training loss and the global gradient norm. 

In a conventional pretraining pipeline, such an event would terminate the run, requiring manual inspection, learning-rate reduction, and a restart from a checkpoint. SOLAR handles this autonomously: the Circuit-Breaker detects the anomaly ($\kappa_{t+1} > 1.5$), aborts the current step, and applies the global Circuit-Breaker penalty to the RL agent. The scheduler then performs a forced PPO update using this failure experience before resuming the model and optimizer from the last safe checkpoint while retaining the updated PPO state.

After the forced PPO update and rollback, the resumed trajectory passes the failure point and completes training without another severe spike. This stress test verifies the automated recovery path under aggressive exploration.

\subsection{Additional Validation on the DeepSeek-V2 3B MoE Model in Megatron}
\label{app:megatron_3b_moe}

\begin{figure}[H]
    \centering
    \begin{subfigure}{0.48\linewidth}
        \centering
        \includegraphics[width=\linewidth]{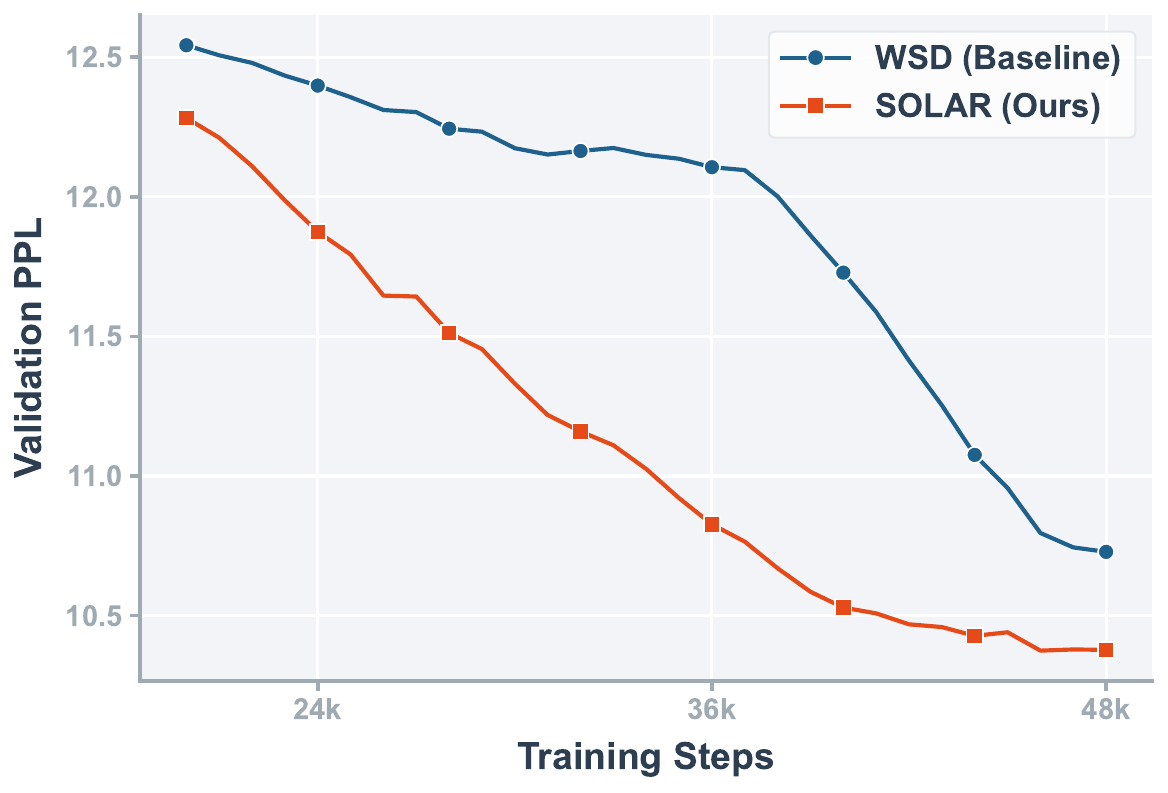}
        \caption{\small Validation PPL}
    \end{subfigure}
    \begin{subfigure}{0.48\linewidth}
        \centering
        \includegraphics[width=\linewidth]{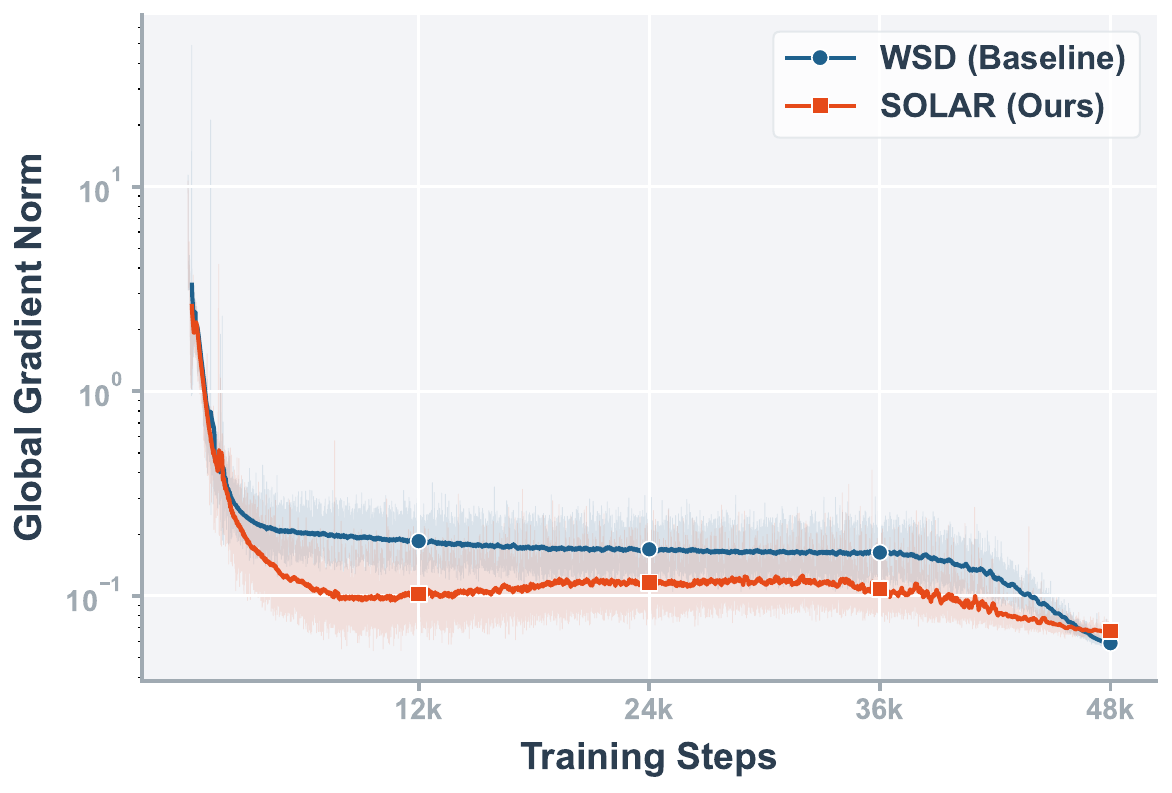}
        \caption{\small Global grad norm}
    \end{subfigure}
    \caption{\small
    Additional validation on the DeepSeek-V2 3B MoE model in Megatron. SOLAR is compared against a matched WSD baseline under the same model, data, tokenizer, optimizer, and training setup.
    }
    \label{fig:megatron_3b_moe}
\end{figure}

We run a separate 3B-scale MoE experiment following the DeepSeek-V2~\citep{deepseekai2024deepseekv2strongeconomicalefficient} architecture in Megatron~\citep{shoeybi2020megatronlmtrainingmultibillionparameter}. This setting changes the model family, corpus, sequence length, distributed stack, and hardware relative to the main dense study.

The model uses sandwich normalization, multi-latent attention, 12 Transformer layers, hidden size 1280, 64 experts, and top-6 routing. The first layer is dense, and the remaining 11 layers use MoE blocks. We train in bf16 with sequence length 4096, global batch size 1024, and 48K updates, or approximately 201B token positions. The corpus is Dolma 3 Mix~\citep{olmo2026olmo3}. Both methods use the same corpus, tokenizer, model configuration, optimizer, token budget, and $32\times$A800 80GB hardware. The WSD baseline uses 2K warmup updates, 12K decay updates, a grid-searched peak LR of $8.6\times10^{-4}$, and a minimum LR of $7\times10^{-6}$. SOLAR uses this same WSD schedule as its base and updates only the residual controller online.

Figure~\ref{fig:megatron_3b_moe} reports validation PPL and global gradient norm. SOLAR reaches a final PPL of 10.38, compared with 10.73 for WSD, and the observed gap peaks near 1.33 around update 38K. Its gradient-norm trace is also lower and smoother over most of the run. We run each method once in this setting. The comparison tests portability to a different MoE stack but does not estimate run-to-run variance.

\subsection{Control Experiments}
\label{app:control_experiments}

These controls complement the replay analyses in Section~\ref{sec:mechanistic_analysis} and Appendix~\ref{app:open_loop_replays}. They compare SOLAR with state-independent stochastic actions and an untrained fixed-policy controller under the same 130M Llama~2 AdamW setup, base schedule, action-scale warmup, and training budget.

\subsubsection{State-Agnostic Stochastic Scheduler}
\label{app:random_residual}

To evaluate state-conditioned policy learning under the same stochastic residual family, we keep the multiplicative residual mapping but replace the learned state-conditioned action with $a_{t,g} \sim \operatorname{TanhNormal}(0,\sigma)$. We select $\sigma$ from $\{1,2,3,4,5\}$ on the tuning seed and report that configuration's final checkpoint. The base schedule, action-scale warmup, and training setup match SOLAR.

The selected state-agnostic scheduler reaches 25.21 PPL, compared with 22.79 for SOLAR (Table~\ref{tab:control_experiments}).

\subsubsection{Untrained Fixed-Policy Controller}
\label{app:untrained_fixed_policy}

This control keeps the SOLAR scheduler architecture, state inputs, and residual action parameterization, but fixes the policy at its random initialization and performs no PPO updates. The base schedule, action-scale warmup, and training setup match SOLAR.

The untrained fixed-policy controller reaches 25.04 PPL. Target-run policy updates reduce this value to 22.79 with SOLAR-online.

\begin{table}[H]
\centering
\footnotesize
\renewcommand{\arraystretch}{1.15}
\setlength{\tabcolsep}{4pt}
\caption{\small Controls for state-independent stochasticity and untrained policy initialization. All experiments use the 130M Llama~2 model with AdamW on C4.}
\label{tab:control_experiments}
\begin{tabular*}{\linewidth}{@{\extracolsep{\fill}}lccccc}
\toprule
\textbf{Method} &
\makecell{\textbf{State-}\\\textbf{conditioned}} &
\makecell{\textbf{PPO}\\\textbf{updates}} &
\makecell{\textbf{Stochastic}\\\textbf{sampling}} &
\makecell{\textbf{Action}\\\textbf{noise}} &
\makecell{\textbf{Final Eval}\\\textbf{PPL} $\downarrow$} \\
\midrule
SOLAR & $\checkmark$ & $\checkmark$ & $\checkmark$ & learned & 22.79 \\[8pt]
\makecell[l]{State-agnostic\\stochastic scheduler} & -- & -- & $\checkmark$ & selected $\sigma$ & 25.21 \\[8pt]
\makecell[l]{Untrained fixed-policy\\controller} & $\checkmark$ & -- & $\checkmark$ & fixed init & 25.04 \\
\bottomrule
\end{tabular*}
\end{table}
The state-agnostic and untrained fixed-policy controls finish at 25.21 and 25.04 PPL, while SOLAR finishes at 22.79.

\subsection{Runtime Accounting}
\label{app:runtime_overhead}

\begin{table}[H]
\centering
\caption{\small Steady-state step-time overhead on the 1B AdamW setting with $8\times$RTX 4090. The post-warmup benchmark window excludes evaluation and checkpointing.}
\label{tab:step_overhead}
\begin{tabular}{lccc}
\toprule
Method & Step Time (s) & Throughput (tok/s) & Overhead Increase (\%) \\
\midrule
Baseline LRS & 1.976 & 66329.01 & 0.00 \\
SOLAR (frozen) & 1.992 & 65799.20 & 0.81 \\
SOLAR (online) & 2.001 & 65499.35 & 1.27 \\
\bottomrule
\end{tabular}
\end{table}

\begin{table}[H]
\centering
\small
\setlength{\tabcolsep}{4pt}
\caption{\small Full-run wall-clock accounting. These measurements include evaluation, checkpointing, state construction, policy inference, action application, PPO updates for the online mode, and any rollback time. No listed run triggered rollback. Dense and Qwen2-MoE runs use $8\times$RTX 4090; the 3B MoE runs use $32\times$A800 80GB.}
\label{tab:runtime_overhead}
\begin{tabular}{lrrrr}
\toprule
Setting & Baseline (h) & SOLAR-online (h) & Online OH & Frozen OH \\
\midrule
1B AdamW & 56.75 & 57.45 & 1.23\% & 0.76\% \\
1B Muon & 65.34 & 66.04 & 1.07\% & 0.69\% \\
Qwen2-MoE 1B & 127.38 & 129.27 & 1.48\% & -- \\
DeepSeek-V2-style MoE 3B & 163.38 & 165.00 & 0.99\% & -- \\
\bottomrule
\end{tabular}
\end{table}

SOLAR does not add an LLM forward or backward pass. Its incremental work comprises reduction-based state construction over resident tensors, a small rank-0 policy network, one scalar-action broadcast per controlled group, and periodic PPO updates in online mode. Freezing the residual policy removes the PPO update while retaining state construction and policy inference.

\paragraph{Policy-side complexity.}
Let $G$ be the number of controlled groups, $D$ the state dimension, $H$ the hidden width, and $A$ the action dimension. The two-hidden-layer policy costs
\[
    \mathcal{O}\!\left(DH + H^2 + HA\right).
\]
per group and
\[
    \mathcal{O}\!\left(G(DH + H^2 + HA)\right).
\]
Here $A=1$ per group. State extraction uses reductions over existing parameters and gradients; PPO backpropagation is amortized because it occurs periodically rather than at every model update.

\paragraph{Step time and full-run wall-clock time.}
Table~\ref{tab:step_overhead} isolates the incremental cost in a steady-state benchmark window. It deliberately excludes evaluation and checkpointing so that the step-level mechanism can be measured without cadence effects. Throughput is computed as
\[
    \operatorname{Throughput}
    = \frac{B_{\mathrm{global}} \cdot L_{\max}}{t_{\mathrm{step}}}.
\]
On 1B AdamW, the measured step-time increases are 1.27\% online and 0.81\% frozen.

Table~\ref{tab:runtime_overhead} times each complete training job from start to finish. It includes evaluation and checkpointing as shared work, as well as SOLAR's checkpoint maintenance and any recovery time. The 1B AdamW full-run overhead is 1.23\%, slightly below the 1.27\% step-time value because common evaluation and checkpoint work enlarge the full-run denominator. Across the listed complete runs, online overhead ranges from 0.99\% to 1.48\%.

The 3B MoE, 1B dense, and Qwen2-MoE settings use 155, 219, and 593 controlled groups and show 0.99\%, 1.23\%, and 1.48\% full-run overhead, respectively. The dense implementation groups one trainable tensor at a time, while the MoE counts follow each codebase's module partition.

\end{document}